\documentclass[sigconf]{acmart}

\AtBeginDocument{%
  \providecommand\BibTeX{{%
    \normalfont B\kern-0.5em{\scshape i\kern-0.25em b}\kern-0.8em\TeX}}}

\usepackage{multirow}
\usepackage{graphicx}
\usepackage{subfig}

\usepackage{textcomp,booktabs}
\usepackage{colortbl}
\definecolor{mygray}{gray}{.9}
\definecolor{mypink}{rgb}{.99,.91,.95}
\definecolor{mycyan}{cmyk}{.3,0,0,0}

\copyrightyear{2020}
\acmYear{2020}
\setcopyright{acmcopyright}\acmConference[MM '20]{Proceedings of the 28th ACM International Conference on Multimedia}{October 12--16, 2020}{Seattle, WA, USA}
\acmBooktitle{Proceedings of the 28th ACM International Conference on Multimedia (MM '20), October 12--16, 2020, Seattle, WA, USA}
\acmPrice{15.00}
\acmDOI{10.1145/3394171.3414034}
\acmISBN{978-1-4503-7988-5/20/10}

\begin{document}
\fancyhead{}

\title{Sharp Multiple Instance Learning for DeepFake Video Detection}

\author{Xiaodan Li}
\authornote{Both authors contributed equally to this research.}
\affiliation{%
  \institution{Alibaba Group, China}
}

\author{Yining Lang}
\authornotemark[1]
\affiliation{%
  \institution{Alibaba Group, China}
}

\author{Yuefeng Chen}
\authornote{Yuefeng Chen is the corresponding author (yuefeng.chenyf@alibaba-inc.com).}
\affiliation{%
  \institution{Alibaba Group, China}}
  
\author{Xiaofeng Mao}
\affiliation{%
  \institution{Alibaba Group, China}
}

\author{Yuan He}
\affiliation{%
  \institution{Alibaba Group, China}
}

\author{Shuhui Wang}
\affiliation{%
  \institution{Inst. of Comput. Tech., CAS, China}
}

\author{Hui Xue}
\affiliation{%
  \institution{Alibaba Group, China}
}

\author{Quan Lu}
\affiliation{%
  \institution{Alibaba Group, China}
}

\begin{abstract}
With the rapid development of facial manipulation techniques, face forgery has received considerable attention in multimedia and computer vision community due to security concerns. 
Existing methods are mostly designed for single-frame detection trained with precise image-level labels or for video-level prediction by only modeling the inter-frame inconsistency, leaving potential high risks for DeepFake attackers.
In this paper, we introduce a new problem of partial face attack in DeepFake video, where only video-level labels are provided but not all the faces in the fake videos are manipulated.
We address this problem by multiple instance learning framework, treating faces and input video as instances and bag respectively.
A sharp MIL (S-MIL) is proposed which builds direct mapping from instance embeddings to bag prediction, rather than from instance embeddings to instance prediction and then to bag prediction in traditional MIL. Theoretical analysis proves that the gradient vanishing in traditional MIL is relieved in S-MIL. To generate instances that can accurately incorporate the partially manipulated faces, spatial-temporal encoded instance is designed to fully model the intra-frame and inter-frame inconsistency, which further helps to promote the detection performance.
We also construct a new dataset FFPMS for partially attacked DeepFake video detection, which can benefit the evaluation of different methods at both frame and video levels.
Experiments on FFPMS and the widely used DFDC dataset verify that S-MIL is superior to other counterparts for partially attacked DeepFake video detection.
In addition, S-MIL can also be adapted to traditional DeepFake image detection tasks and achieve state-of-the-art performance on single-frame datasets. 

\end{abstract}

\ccsdesc[500]{Computing methodologies~Computer vision problem}
%
\keywords{DeepFake; Multi-Instance Learning; Weighting; Temporal}

\begin{teaserfigure}
  \includegraphics[width=\textwidth]{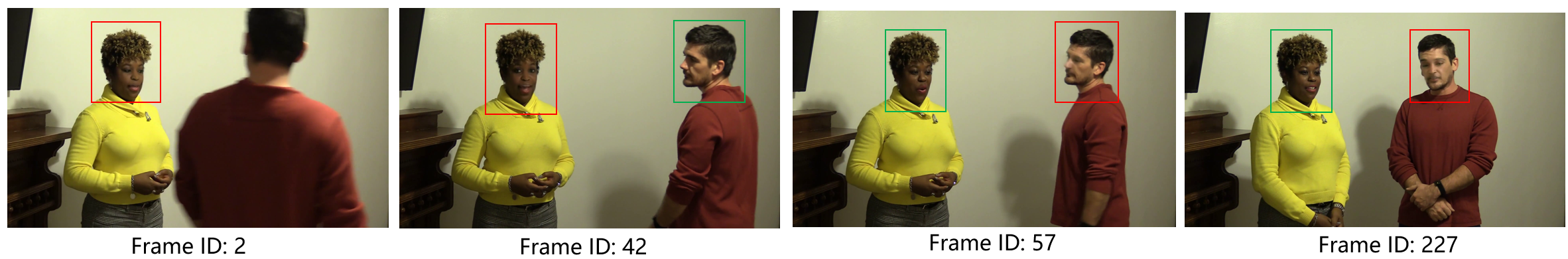}
  \vspace{-7mm}
  \caption{
Example of partially attacked DeepFake video. The green and red boxes represent real and fake faces respectively. This figure illustrates that not all faces in a fake video are manipulated. Real and fake faces may appear in the same frame. Face labels of one person in nearby frames may also be different.
  }
  \vspace{3mm}
  \label{fig:teaser}
\end{teaserfigure}
\vspace{-5mm}

\maketitle

\section{Introduction}
The rise of face manipulation techniques has fueled the advent of forgery face images and videos~\cite{DeepFakes,FaceSwap,Justus2018Face2Face,Deferred}. 
As the swapped faces become more and more lifelike with advanced deep learning models such as Generative Adversarial Networks~(GAN), these techniques, {\it e.g.}, DeepFake~\cite{DeepFakes}, are abused for malicious purposes, {\it e.g.}, face recognition attack or even political persecution. To enhance the content security, it is crucial to develop more advanced forgery detection technology.

In previous study, the DeepFake video detection is either treated as a forgery image detection problem, or addressed as a video level fake prediction problem with video feature learning.
Different from traditional setting, we address a new problem that exists more ubiquitously in real situations, {\it i.e.}, partial faces attack in a video where not all faces in a fake video are manipulated. This problem can be illustrated from two aspects,
as shown in Fig.~\ref{fig:teaser}. Firstly, real and fake faces may occur simultaneously in one frame. Secondly, faces of one target person may be partially manipulated in one DeepFake video due to arbitrary content attack behaviors. 

Existing DeepFake video detection methods can be divided into frame-based~\cite{DingSwapped,Shahroz,Marra2018Detection,Faceforensics,Faceforensics++} and video-based methods~\cite{sabir2019recurrent}.
Frame-based methods meet with three major difficulties when applied to fake video detection with partial faces attack. First, they are generally data hungry and require a remarkable number of labeled frames for training~\cite{X-ray,Guera}.
Their performances may drop significantly when detecting partially attacked DeepFake videos since they simply treat video labels as frame labels.
Besides, every sampled frame needs to be predicted for final prediction during inference, which is very computationally exhaustive.
The video-level decision making based on frame-level representation is another key issue for frame-based methods. 
Using the maximum fake scores over faces or averaging them may lead to unexpected false alarm or false negative predictions.
For video-based methods~\cite{sabir2019recurrent}, more attention is paid on the temporal feature modeling, which is less effective for DeepFake detection under partial attack situation since it is also highly dependent on appearance modeling. Thus, traditional solutions leave potential high risks for DeepFake attackers.

Considering that videos are only given the video-level annotation in many practical situations, we address DeepFake video detection based on multiple instance learning~(MIL) framework by treating faces and input video as instances and bag respectively. One bag consists of multiple instance and only bag labels are available. When one instance is positive, the corresponding bag will be labeled as positive. This setting is naturally suitable for DeepFake video detection task.
However, in traditional MIL, instance predictions have to be learned based on bag labels, and bag prediction has to be made based upon instance predictions. Thus, gradient vanishing occurs constantly when the fake score of one instance in a bag is high. This issue leads to partial fake instance detection results~\cite{ilse2018attention}. 
Research endeavor has been dedicated to more careful network design~\cite{wang2018revisiting,ilse2018attention}. Nevertheless, theoretic insight on how the gradient vanishing problem can be alleviated has not been provided. The research challenge can be further analysed from the attacker side where DeepFake \cite{DeepFakes} tampers video on specific frames, which inevitably causes spatial-temporal inconsistency in nearby frames, evidenced as the content jitters. 
However, in traditional MIL, instances are generated by considering appearance cues only. This issue has to be tackled with a careful design of instance structure by considering both the rich spatial and temporal cues in videos.


In this paper, we propose Sharp Multiple Instance Learning~(S-MIL) for DeepFake video detection.
Different from traditional MIL where bag prediction has to be made upon instance prediction built on the multi-dimensional instance feature space, we apply Sigmoid on a weighted sum operation on a bag of instance embeddings to directly produce the bag prediction. 
Consequently, the loss on bag prediction can be directly propagated to instance embeddings, which can help to ease the gradient vanishing problem. Besides, the proposed S-MIL can still be explained in traditional MIL probabilistic framework, which benefits for comprehensive theoretic analysis. Accordingly, it can be theoretically derived that the gradient surface of the loss with respect to the instance prediction of S-MIL tends to be much sharper than traditional MIL, so the gradient vanishing problem during the back propagation from bag prediction to instance embeddings can be effectively alleviated.
Based on the proposed S-MIL, we propose to produce multiple spatial-temporal instances to fully model the intra-frame and inter-frame inconsistency caused by independent frame attacking behavior to generate forgery faces.

To inspire research on video-level DeepFake detection under partial faces manipulation situations, we construct a new dataset named FaceForensics Plus with Mixing samples~(FFPMS) for partially attacked DeepFake video detection. 
The dataset contains both frame-level and video-level annotations for more comprehensive evaluation while models are allowed to be trained using only the video-level annotations. Experiments on FFPMS dataset verify that S-MIL is superior to other competitors for partially attacked DeepFake video detection. 
Also, S-MIL can make a more accurate judgement on whether a video is fake or not than other methods on DFDC benchmark. S-MIL can also be adapted to traditional single-frame detection task and achieve state-of-the-art performance on the FF++ datasets \cite{Faceforensics++}, showing a promising result for face information protection. 

The major contributions of our paper include: 

\begin{itemize}
\item 
We introduce a new problem named partial faces attack in DeepFake video detection and propose S-MIL to address this problem. 
To the best of our knowledge, this is the first MIL-based solution for DeepFake video detection.
Theoretical analysis shows that S-MIL can alleviate the gradient vanishing problem effectively.

\item We design a new spatial-temporal instance to capture the inconsistency between faces, which can help to improve the accuracy of DeepFake detection.

\item We propose FFPMS dataset with both frame-level and video-level annotations for video-based DeepFake face detection.

\item We verify that our approach is superior to other counterparts for video face forgery detection and our S-MIL can also be adapted to traditional single-frame detection.
\end{itemize}

\section{Related Work}
DeepFake detection obtains considerable attention due to the security concerns caused by the abuse of face swapping techniques. In this section, we briefly review typical detection methods and works related to multi-instance learning.

\subsection{DeepFake Detection} 
According to the detection target, DeepFake detection can be divided into image-level detection and video-level detection. Early works~\cite{Exposing,Rich,CFA-aware} achieve the forgery face detection by handcraft features, as the face manipulation techniques are limited at that time. These traditional manipulation methods typically swap the face area based on the facial landmarks and then utilize some post-processing techniques to make the boundary of swapped face inconspicuous.

With the development of generative adversarial network (GAN)~\cite{radford2015unsupervised,goodfellow2014generative}, the forgery faces become more and more realistic. Therefore, some works~\cite{CozzolinoRecasting,Rahmouni2017Distinguishing,Bayar2016A,Fakespotter} begin to utilize deep networks to learn discriminative features or find manipulation traces for DeepFake detection.
For instance, Rossler \emph{et al.}~\cite{Faceforensics++} introduce a simple but effective Xception Net as a binary DeepFake image detector and Li \emph{et al.}~\cite{li2019exposing} as well as Lingzhi \emph{et al.}~\cite{X-ray} aim at the finding traces such as blending boundaries left by the DeepFake generation methods with deep neural networks.
They are trained with frame-level labels and can achieve high accuracies for DeepFake image detection.

Compared to the forgery images, video-level forgeries \cite{DeepFakes,Justus2018Face2Face,Deferred} are more harmful, since they look more convincing with real audio. The above frame-based methods \cite{Guera,X-ray,Mesonet,li2019exposing} can also be used to detect video forgeries. The typical solution is choosing some frames randomly from the video and taking the max or average score as the final score of this video for classification (real or fake). However, the maximum operation may lead to a false alarm while averaging may cause false negative prediction because the scores on manipulated faces are overwhelmed by those of normal faces. Previous researchers have also explored to address DeepFake video detection as a video level prediction problem with video feature learning. Likewise, Sabir \emph{et al.}~\cite{sabir2019recurrent} propose to use recurrent convolutional models~\cite{hochreiter1997long} to exploit the temporal information in DeepFake videos. However, they are limited in some cases since face forgery detection is highly dependent on appearance modeling.


\subsection{Multiple Instance Learning} 
For a typical Multiple-instance learning~(MIL) method, the model receives a group of labeled bags which consist of many instances, rather than requiring instances that are labeled individually. MIL aims to learn a model that can predict the bag label~\cite{Keeler1990Integrated,Amores2013Multiple,Revisiting}. 

In the early work of MIL, pre-computed features or hand-craft features are utilized to represent the instances \cite{Dietterich2001Solving,Maron1998}. The boom of deep learning-based approaches further enhances the ability of multi-instance learning by a large margin \cite{Revisiting,Oquab}, especially for medical image scenarios \cite{medical,Sirinukunwattana2016Locality}. 
Generally speaking, MIL algorithms can be divided into two fold: instance-space paradigm and embedded-space paradigm~\cite{wang2018revisiting}. Traditional MIL generally follows instance-space paradigm, which aggregates instances on output layer. Some works claim that it is inflexible and propose to do mean-pooling or max-pooling in embedded-space~\cite{MIML,image-level,Deepzhu}. But both operators are non-trainable and susceptible to extreme values, which potentially limits their applicability.

The attention mechanism can relive the above weakness by focusing on key instances, which is widely used in recent  works~\cite{Show2015,Bahdanau2014,Lin2017}. 
Inspired by the idea of lp-norm pooling~\cite{Feng2011Geometric}, Zagoruyko \emph{et al.}~\cite{ZagoruykoPaying} propose a novel attention mechanism that requires a student model to mimic the feature map of an attention model (teacher model), which enhances the performance of the student model significantly.
Recently, the attention mechanism has been also used in some MIL works~\cite{Explaining, Explicit, ilse2018attention}. For example,  Ilse \emph{et al.} \cite{ilse2018attention} claim that traditional MIL may face the gradient vanishing problem and they propose an attention-based MIL network that achieves the attention weights training by a small neural network. However, theoretic insight on how the gradient vanishing problem can be alleviated has not been provided.


\section{Our Approach}

\begin{figure*}[!t]
 \centering
  \includegraphics[width=1\linewidth]{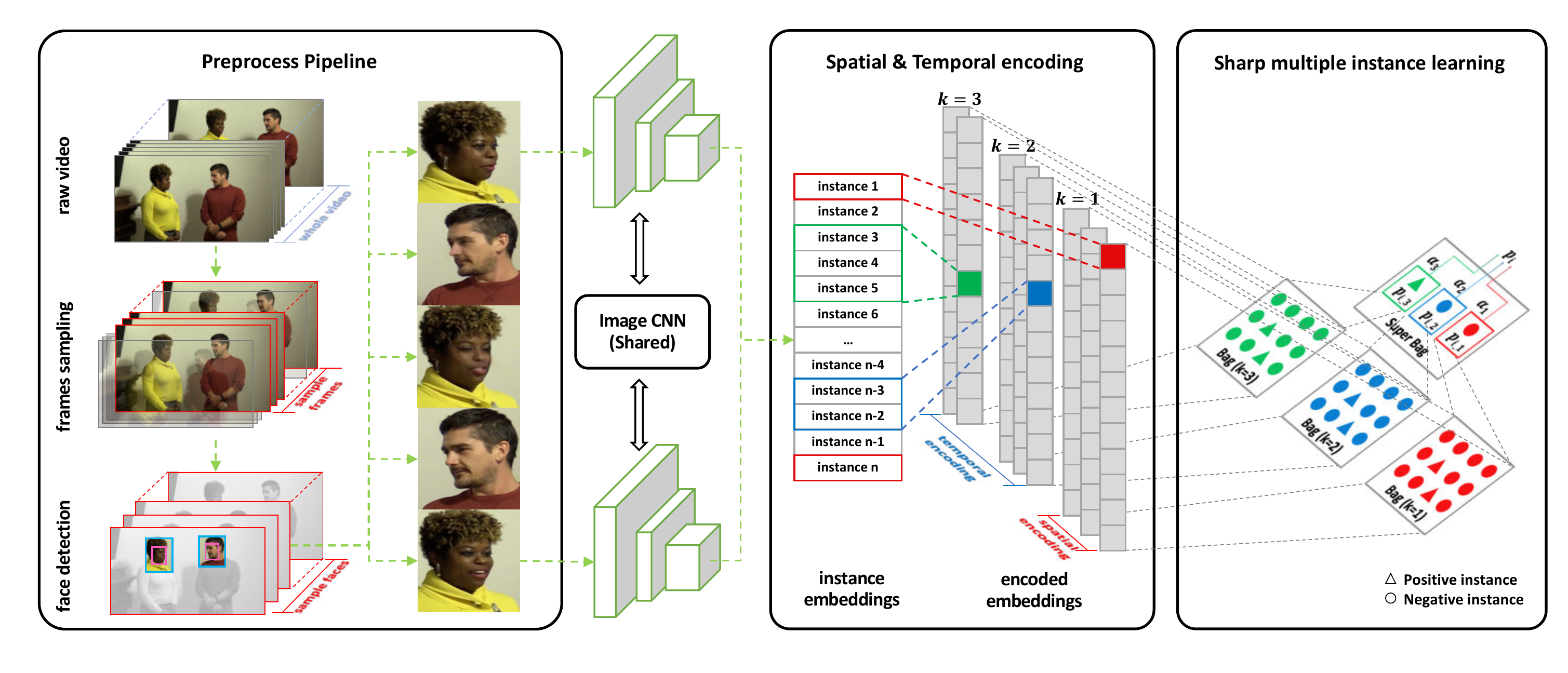}
  \parbox[t]{1.0\textwidth}{\relax}
  \vspace{-5mm}
  \caption{\label{fig:framework}
           Structure of the proposed algorithm which consists of three modules. Given a frame sequence from a video, our approach firstly detects faces and extracts individual feature maps by CNN. Then, the feature maps of the face sequences are encoded as spatial-temporal instances to get multiple spatial-temporal bags. After that, the resulted features of instances in different bags are integrated by the proposed sharp multi-instance learning method. Finally, S-MIL is performed on these bags to get the final video prediction~(real or fake).
}
\end{figure*}

In this section, we illustrate the proposed algorithm in detail.
As shown in Fig.~\ref{fig:framework}, our algorithm consists of three key components.
First, face detection is conducted on the sampled frames in input videos. The extracted faces are then fed into a CNN to obtain features as instances.
Second, spatial and temporal instances are extracted with corresponding encoding branches to form spatial-temporal bags with different temporal kernel sizes. These bags are used together to represent a video.
Last, S-MIL is performed on these bags to get the final fake scores of all bags,  which can derive a final fake score for the whole video.

\subsection{Problem Formulation}
Formally, for $N$ videos in an input batch, the goal for binary classification is to minimize the following objective function:
\begin{equation}
\mathcal{L} = -\sum_{i=1}^{N} (y_{i}\text{log}p_{i} + (1-y_{i})\text{log}(1-p_{i})), \quad p_{i} = F(W, x_{i}),
\end{equation}
where $y_i$, $p_{i}$ are the label, prediction of the $i$-th video $x_i$. $F$ and $W$ are the prediction model and trainable parameters respectively. 

For one single video, 
traditional DeepFake detection methods tend to convert the above video level prediction task to frame level task by training a supervised per-frame binary classifier~\cite{Faceforensics++, Mesonet, X-ray}. During inference, averaging or maximizing is performed to get the final prediction of input video:
\begin{equation}
p_{i} =
\left\{
             \begin{array}{lr}
             \frac{1}{M}\sum_{j=1}^{M}p^{j}_{i} = \frac{1}{M}\sum_{j=1}^{M}F(W, x^j_i), \quad \text{Average}\\
             \max \limits_{j \in [1,M]}p^{j}_{i} = \max \limits_{j \in [1,M]}F(W, x^j_i), \quad \text{Maximum}
             \end{array}
\right.
\end{equation}
where $p^j_i$ and $M$ are the fake score of the $j$-th frame and the total frame number in video $x_i$ respectively.

The above formulation will force every frame in fake videos to be predicted as fake. However, it is inaccurate since fake videos may only contain part of the manipulated target faces. 
Simply applying the video label to frame label may result in noises.
Furthermore, when generating video-level label with face labels, using the maximum fake scores may lead to false alarms while averaging may lead to false negative predictions.
Under MIL framework, we treat videos as an integral instead of analyzing frame by frame.

For a typical MIL work, 
the model receives labeled bags which consist of different number of instances, rather than requiring instances labeled individually. MIL classifier aims to predict the labels of testing bags. A bag is labeled as negative when all the instances in this bag are negative, while a bag is labeled as positive when there is at least one instance in this bag that is positive ~\cite{Revisiting}. Formally, for a given bag with a bag label $y_i \in \left \{0, 1\right\}$, it consists of $M$ instances. For the $i$-th bag in MIL,
\begin{equation}
y_i =
\left\{
             \begin{array}{lr}
             0, \quad \text{if} \,\,\, \sum_{j=1}^{M} y^{j}_i= 0  \\
             1, \quad \text{otherwise} 
             \end{array}
\right.
\end{equation}

We fit MIL into DeepFake video detection.
Following the traditional MIL definition, the probability for a bag
is:
\begin{equation}
p_{i} = 1-\prod_{j=1}^M(1-p^{j}_{i}).
\end{equation}
Then the objective function of input bags can be denoted as:
\begin{equation}
\mathcal{L} = -\frac{1}{N}\sum_{i=1}^{N} (y_{i}\text{log}(1-\prod_{j=1}^M(1-p^{j}_{i})) + (1-y_{i})\text{log}(\prod_{j=1}^M(1-p^{j}_{i}))).
\end{equation}

For instances in positive bags, the gradient is:
\begin{equation}
\frac{\partial \mathcal{L}}{\partial p^j_i} = \frac{\partial \mathcal{L}}{\partial p_i} \frac{\partial p_i}{\partial p^j_i} = \frac{\partial \mathcal{L}}{\partial p_i} (\prod_{k=1, k\ne j}^M(1-p^{k}_{i})).
\end{equation}

When there is one instance predicted as positive, other positive instances will face the gradient vanishing issue. This will limit its applications such as judging one frame is fake or not.

\subsection{Sharp Multi-Instance Learning}
Generally speaking, MIL algorithms can be divided into two folds: instance-space
paradigm and embedded-space paradigm~\cite{wang2018revisiting}.
Traditional MIL generally follows instance-space paradigm, which aggregates instances on label output layer. Since the individual labels are unknown, there is a bottleneck for the prediction accuracy, caused by the insufficient training of instance-level model~\cite{wang2018revisiting}.
To integrate instances in embedded-space, we propose to use a simple sum operator to fuse features of different instances.
Let $H_i = [h^1_i, . . . , h^M_i]$ be a bag of $M$ embeddings extracted by the backbone network, and the classifier layer be $W$. Denote the fake scores for the $i$-th video and the $j$-th frame in the $i$-th video as $p_i$ and $p_i^j$, respectively, then:
\begin{equation}
\begin{aligned}
p_{i} = \frac{1}{1+e^{-W\sum_{j=1}^M(h^j_i)}}, \quad p^{j}_{i} = \frac{1}{1+e^{-W h^j_i}}. \label{eq.prob}
\end{aligned}
\end{equation}
Actually, our proposed MIL can also be formulated in tradition MIL manner as in Eq.~\ref{eq:S-MIL}, which is helpful in analyzing its theoretic merits.

\begin{equation}
\begin{aligned}
p_{i} = \frac{1}{1+e^{-W\sum_{j=1}^M(h^j_i)}}
        = \frac{1}{1+\prod_{j=1}^M(\frac{1}{p^j_i}-1)}. \label{eq:S-MIL}
\end{aligned}
\end{equation}

\noindent\textbf{The sharp characteristic}
With Eq.~\ref{eq:S-MIL}, it can be found that the proposed MIL can ease the gradient vanishing problem, which is proved in the supplementary. Besides, for an intuitive explanation, we set $M$ to 2 to show the gradient surface ${\partial \mathcal{L}}/{\partial p^j_i}$ in 3D space with respect to different $p^1_i$ and $p^2_i$. As shown in the supplementary, the gradient surface of the proposed MIL looks sharper than traditional MIL, resulting in a smaller area with vanished gradients. This validates that the proposed formulation in S-MIL can relieve the gradient vanishing problem.

\noindent\textbf{Weighting mechanism}
Eq.\ref{eq:S-MIL} simply treats instances equally without any focus, which tends to be sub-optimal. For example, if $[p^1_i, p^2_i, p^3_i] = [0.1, 0.1, 0.9]$, with Eq.~\ref{eq:S-MIL}, $p_i < 0.5$, which is not consistent of the definition of MIL. 

Referring to the ideas of boosting~\cite{chen2016xgboost} and focal loss~\cite{lin2017focal}, which are designed to emphasizing the learning on hard examples, we propose to ``boost'' the informative instances in bags.
To the end, the proposed S-MIL is defined as:
\begin{equation}
p_{i} = \frac{1}{1+\prod_{j=1}^M(\frac{1}{p^j_i}-1)^{\alpha^j_i}} \label{eq.ASMIL}, \quad a^{j}_i=\dfrac{\text{exp}\{{\mathbf{w}^\top h^{j}_i}\}}{\sum_{j=1}^{M}\text{exp}\{{\mathbf{w}^\top h^{j}_i}\}},
\end{equation}
where $\mathbf{w}$ is the learnable parameter of a neural network.
It can adjust the weights of different instances within one bag. The proposed S-MIL can also be converted to embedded-space and it is a weighted sum of instances represented by low-dimensional embeddings $h^j_i$: 
\begin{equation}
\begin{aligned}
p_{i} = \frac{1}{1+\prod_{j=1}^M(\frac{1}{p^j_i}-1)^{\alpha^j_i}}
    = \frac{1}{1+e^{-W\sum_{j=1}^M(\alpha^j_i h^j_i)}}.
\end{aligned}
\end{equation}

There are some advantages of the weighting mechanism. First, it is designed to enhance the weights of informative instances to extract discriminative representations for input bags.
Second, for positive bags, the key instances are mostly positive ones, which should be assigned with high weights. Thus, it can help to interpret the final decision. 

In similar research routine, attention-based multi-instance learning has been studied by previous work~\cite{ilse2018attention}, which claims traditional MIL may face the gradient vanishing problem. 
Nevertheless, the attention-based approaches explore the instance-bag relation in a data-driven manner, and theoretic insight on how the gradient vanishing problem can be alleviated has not been provided.
In comparison, our method can be naturally explained with traditional MIL probabilistic framework, and also has the advantage of attention-based models where the gradients can be back-propagated to instance embeddings without serious vanishing. 


\begin{figure}[t]
 \centering
  \includegraphics[width=1.0\linewidth]{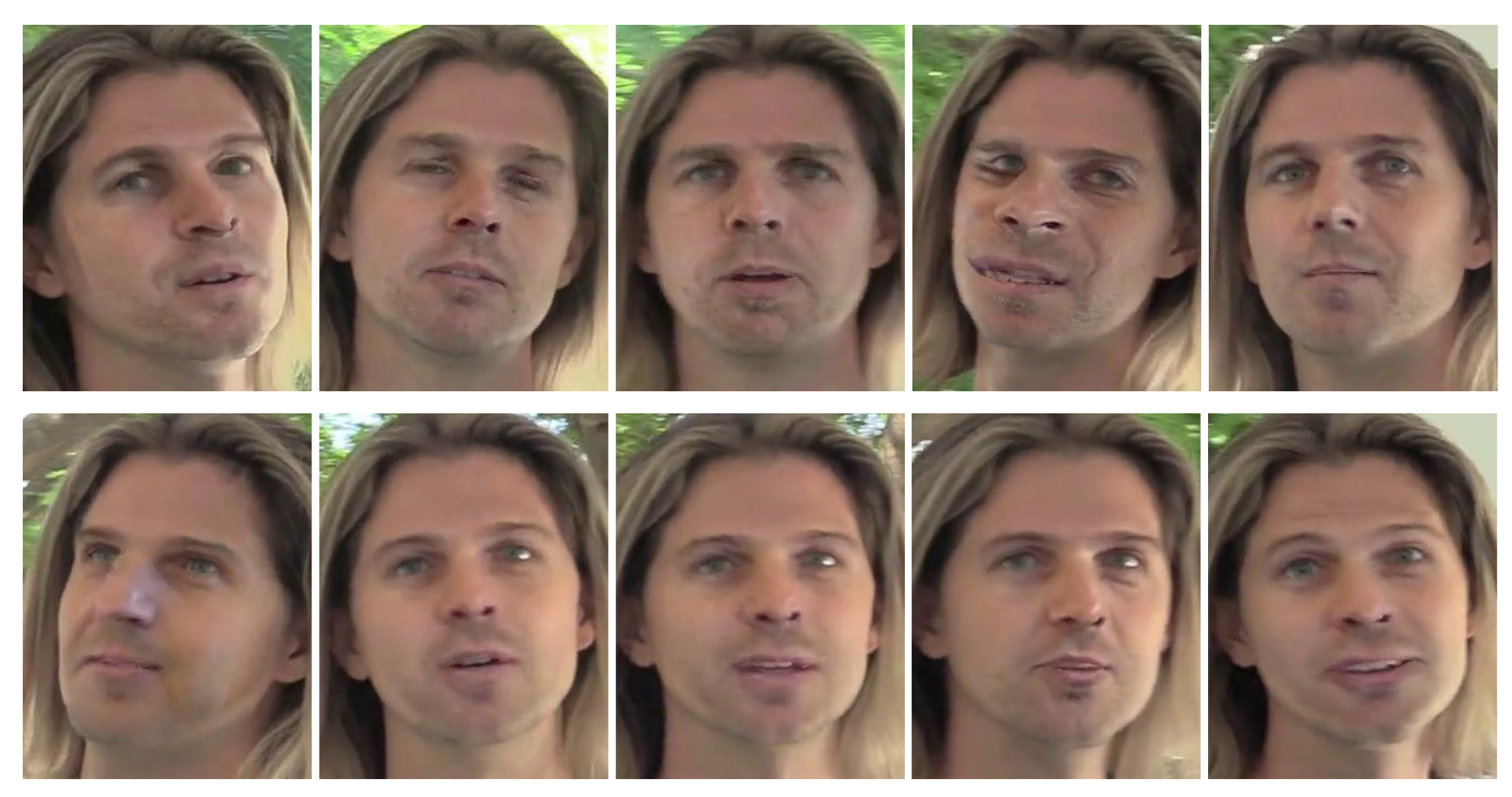}
  \vspace{-2mm}
  \caption{\label{fig:temporal}
           Visualization for manipulated faces across frame sequences. Faces vary a lot along the temporal dimension.
}
\end{figure}

\subsection{Spatial-temporal Instances}
The S-MIL and traditional MIL are developed based on instances or frames that are independent to each other and they are organized at the spatial-level. It is true that faces are manipulated individually and they are irrelevant to each other. However, from another point of view, since they are made separately, the sequences along the temporal dimension may not be as smooth as real face sequences.

As described in Fig.~\ref{fig:temporal}, the forgery faces vary drastically along the temporal dimension. This important cue can be utilized for DeepFake video detection.
To model inter-frame consistencies between faces extracted from nearby frames, based on the proposed S-MIL framework, we design a new spatial-temporal instance by adding a spatial-temporal branch with multiple temporal kernels. Inspired by the previous sentence classification method \cite{kim2014convolutional}, we utilize the 1-d CNN to encoding input frames in the temporal level. 

Let $Conv1d_{k,r}$, be a 1-d convolutional block, where $r$ and $k$ are the number and size of filters respectively.
Feeding feature map $H_i = [h^1_i, h^2_i, ..., h^M_i]$ generated by the CNN backbone, after zero padding, into $Conv1d_{k,r}$, we produce a $M \times r$ feature map. 
Then ReLU activation function is performed on this feature map for non-linearity.
More formally, we describe the above process as
\begin{equation}
c^{k}_{i} = ReLU(Conv1d_{k,r}(H_i)),
\end{equation}
where $c^{k}_{i}$ is the spatial-temporal encoding bag for $H_i$ with kernel size $k$.

When we set the kernel size $k$ as 2, it allows two adjacent rows in $h^j_i$ to interact with each other. As the kernel size grows, more nearby rows are exploited simultaneously. We employ kernel size $k$ of 2 and 3 for temporal encoding and at the same time, keep $k=1$ to retain the spatial encoding instances as well. The number of filters for each kernel size is set to 512.

Finally, several spatial-temporal encoded bags with different kernel sizes compose a super bag to represent a video, which is processed by S-MIL to get the final fake score for the input video. Denote $p_{i\_k}$ as the fake score of bag encoded with kernel size $k$, the final prediction of the input video is:
\begin{equation}
p_i = \text{S-MIL}(p_{i\_1}, p_{i\_2}, ..., p_{i\_k}), \quad \text{where} \quad p_{i\_k}=\text{S-MIL}(c^k_i).
\end{equation}

\begin{figure*}[t]
 \centering
  \includegraphics[width=0.9\linewidth]{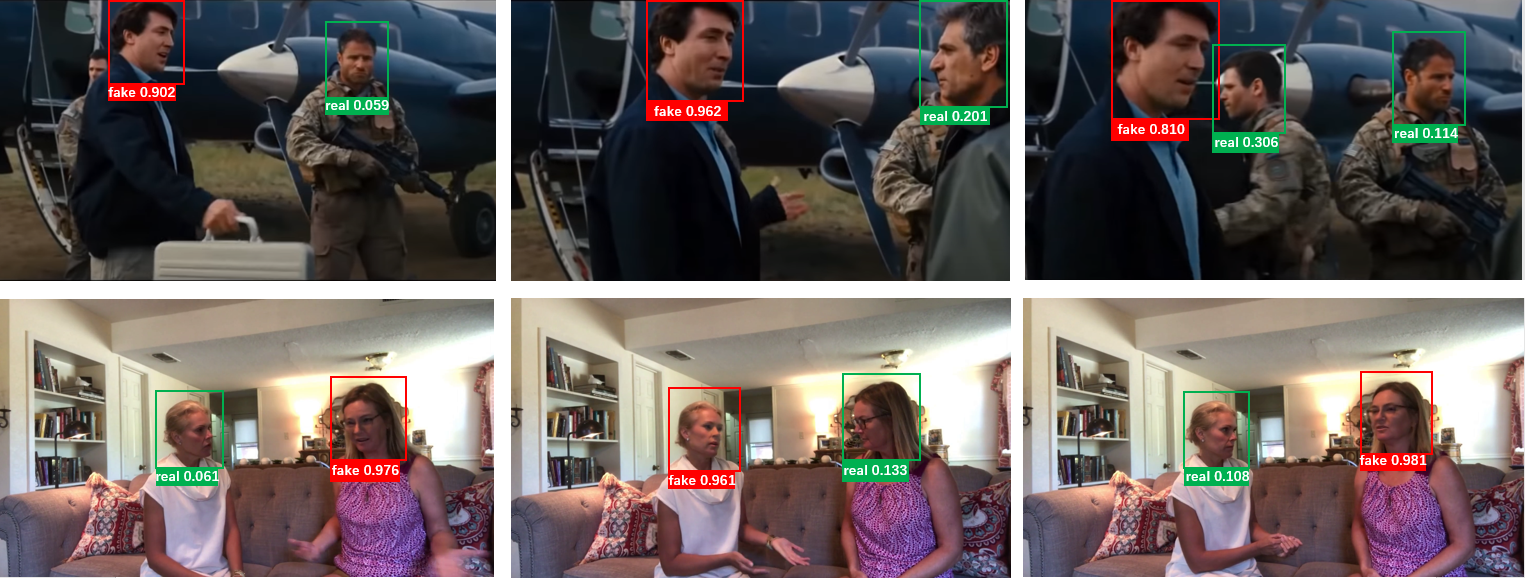}
  \parbox[t]{0.9\textwidth}{\relax}
  \vspace{-2mm}
  \caption{\label{fig:hr}
           Visualization results with S-MIL for DeepFake video detection of two partially attacked example frame sequences~\cite{Partially, DFDChallenge}. 
           Faces in green and red boxes are real and fake respectively and the scores associated with these boxes are the corresponding fake scores predicted by the proposed method.
}
\end{figure*}
\vspace{-2mm}

\subsection{Loss Function}
Since we define the MIL DeepFake video detection as a binary classification task, we choose the Binary Cross Entropy~(BCE) as our loss function. Given the positive probability $p_{i}$ of the $i$-th video, the loss is calculated as follows and the entire network is trained towards minimizing the following loss:
\begin{equation}
\mathcal{L} = -\dfrac{1}{N}\sum_{i=1}^{N}(y_i \cdot \text{log}(p_i)+(1-y_i) \cdot \text{log}(1-p_i)). 
\end{equation}

\section{Experiments}
In this section, we experimentally evaluate the proposed algorithm for DeepFake video detection and compare it with other counterparts in terms of partially attacked datasets evaluation,
as well as evaluations on fully attacked benchmarks under video-level and frame-level settings. 
Also, we conduct an ablation study to explore the influence of proposed components as well as the number of sampled frames during the inference.
For clarify, we denote S-MIL as the version of method with $k=1$, and S-MIL-T as the version with $k=1,2,3$ in subsequent experiments. 
We also provide a demo video in supplementary. 

\subsection{DeepFake Video Detection Datasets}

\noindent\textbf{FaceForensics ++(FF++)}~\cite{Faceforensics++}: It is a recent released benchmark dataset and has been widely used for evaluation in different works~\cite{Faceforensics++, X-ray}.
It consists of 1000 original videos and corresponding fake videos which are generated by four typical manipulation methods: DeepFakes~(DF)~\cite{DeepFakes}, Face2Face~(F2F)~\cite{Justus2018Face2Face}, FaceSwap~(FS)~\cite{FaceSwap}, NeuralTexures~(NT)~\cite{Deferred}.
It provides frame-level labels and nearly every frame and face in fake videos are manipulated.

\noindent\textbf{Celeb-DF}~\cite{li2019celeb}: Celeb-DF dataset is composed of 5639 YouTube videos based on 59 celebrities. These videos are tampered by the DeepFake method and contain more than 2 million frames in total. The quality of the tampered videos looks great with little notable visual artifacts. Nearly every frame in the video is manipulated. 

\noindent\textbf{Deepfake Detection Challenge~(DFDC)}~\cite{dolhansky2019deepfake}: DFDC is a preview dataset released by DeepFake detection challenge$\footnote{https://deepfakedetectionchallenge.ai/}$. This dataset is generated by two kinds of unknown synthesis methods on 1131 original videos. Totally, it creates 4113 forgery videos based on humans of various ethnic, ages and genders. It only provides video-level labels and some fake videos contain many un-manipulated frames and faces.

\noindent\textbf{FaceForensics Plus with Mixing samples~(FFPMS)}: Since existing datasets with frame labels have few video samples with mixed real/fake instances, which is not beneficial for analyzing performances in partial face attack scenario. We construct a new benchmark called FaceForensics Plus with Mixing samples~(FFPMS). 

Specially, we sample 20 frames from each video in the FF++ testing set whose compress rate is c40 and replace several fake frames with real frames from corresponding videos. The number of replaced fake frames ranges from 1 to 19. The resulting dataset has 14000 frames, containing four kinds of fakes~(DF, F2F, FS, NT) and the original ones. With both frame-level labels and video-level labels, this dataset is designed to model fake videos with un-manipulated frames and faces, which is very common in Youtube videos but rare in existing datasets except for DFDC. However, DFDC does not contain frame-level labels.

\subsection{Implementation Detail} 
We adopt XceptionNet as a default backbone network as other works for fairness~\cite{Faceforensics++, X-ray}.
For spatial and temporal encoding, we set the kernel sizes to $1,2,3$.
The number of filters of each kernel size is set to $512$. All frame-based and video-based models are trained with only $20$ uniformly sampled frames in each video. For video-based models, we utilize a batch size of $32$ for training and finish the training process after $30$ epochs. During each training epoch, $8$ frames are randomly extracted from the sampled $20$ frames. Faces are detected and cropped from the resulting frames as the video input. 
We finetune on the ImageNet pretrained xception model with a learning rate of $0.0002$, which is reduced to half every $5$ epochs. During training, we adopt the Adam method ~\cite{kingma2014adam} as the optimizer. Only random crop and horizontal flip are employed during training since we focus more on network designing. By default, $20$ frames of each  video are uniformly extracted as network inputs for efficiency during testing. 
More details about face detectors can be found in supplementary.

\begin{figure*}[t]
 \centering
  \includegraphics[width=1\linewidth]{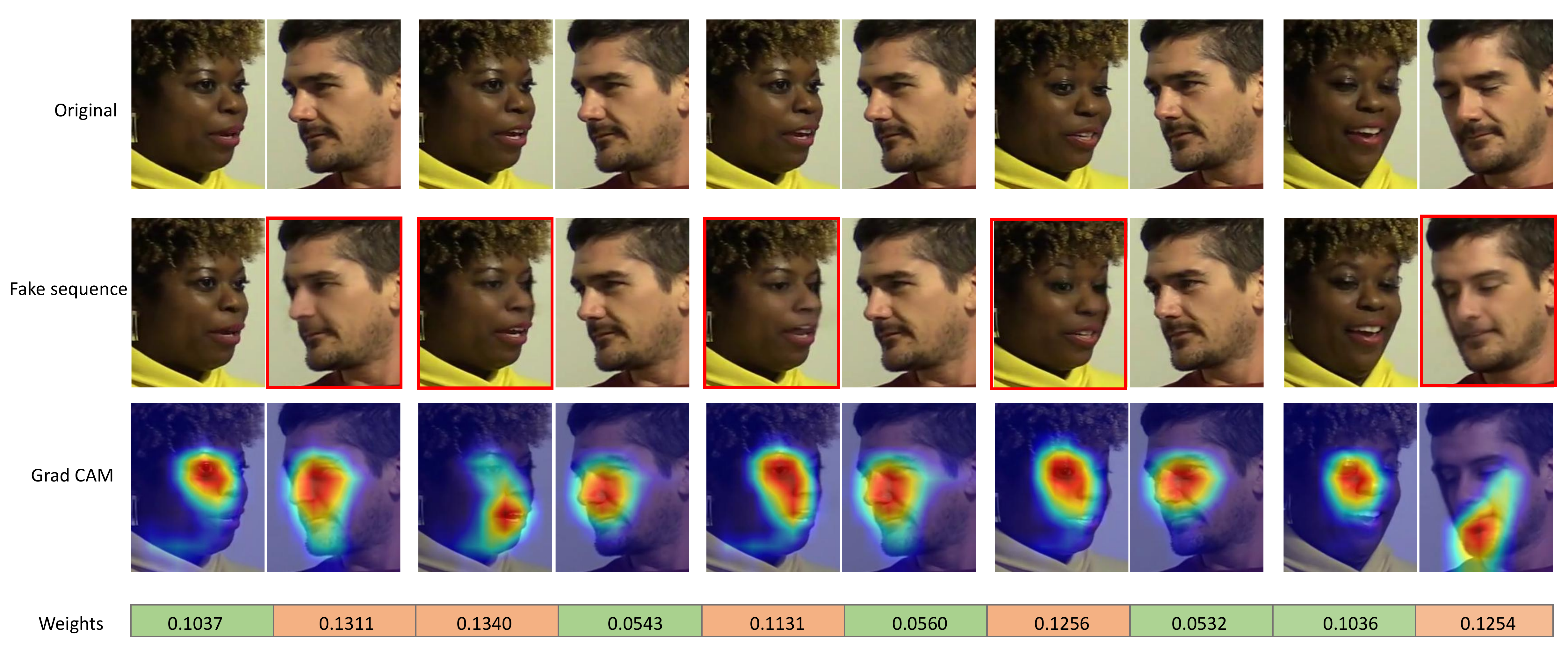}
  \parbox[t]{1.0\textwidth}{\relax}
  \vspace{-6mm}
  \caption{\label{fig:gradcam}
          Visualization of a frame sequence and the corresponding heatmaps extracted with gradcam~\cite{selvaraju2017grad}. Faces in red bounding boxes are fake. The last row shows the weights of each face. Forgery faces have higher weights than real faces.
}
  \vspace{-2mm}
\end{figure*}

\subsection{Evaluation on Partially Attacked Datasets}
We conduct the experiment on DFDC, Celeb, and FFPMS datasets to evaluate the generalization ability of S-MIL.  We compare the proposed S-MIL with frame-based model such as XceptionNet~\cite{Faceforensics++}, D-FWA~\cite{li2019exposing} and some video-based models such as LSTM~\cite{hochreiter1997long} and I3D~\cite{carreira2017quo}. Since the FF++ dataset, from which FFPMS is extracted, contains few 
unmodified faces in fake training videos, we replace 25\% of the whole fake frames with their corresponding real ones in the fake videos for training the S-MIL model. 
The experiment is evaluated by the average accuracy of fake and real testing videos.

\begin{figure}[t]
  \centering
  \small
  \includegraphics[width=0.8\linewidth]{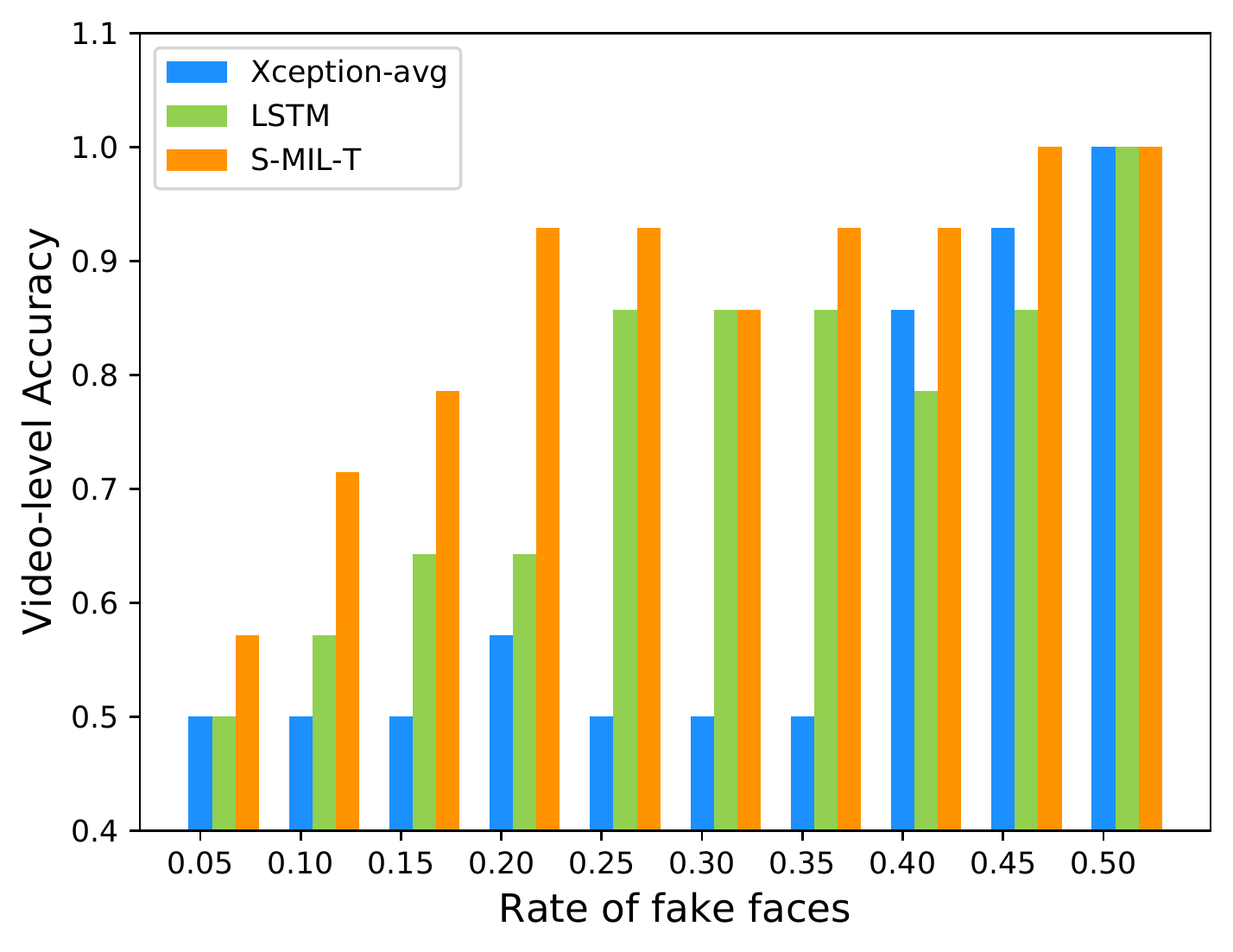}
  \caption{\label{fig:noise}
           Comparison for S-MIL-T, XceptionNet-avg, LSTM on different rate of fake faces in input sequences of FFPMS dataset. The S-MIL-T can get a good performance even when the fake rate is low while others can not.
}
\end{figure}

For baseline XceptionNet, which is a frame-based method, there are two ways for converting it to a video-level model, {\it i.e.}, averaging fake score of each frame or treating the maximum fake score as the video-level prediction.
For thorough analysis, we experiment on both cases.
Table~\ref{tbl:FFPMS} summarizes the accuracy of different detectors with respect to each type of manipulated video in different datasets.
As shown in Table~\ref{tbl:FFPMS}, the proposed S-MIL performs better than both frame-based and video-based methods in most cases. Besides, either in max or average ways, XceptionNet performs much worse than the proposed S-MIL.
When equipped with spatial-temporal encoding as in S-MIL-T, the proposed method gets a better performance. This validates the effectiveness of the proposed S-MIL and spatial-temporal instances.
We also compare the accuracies under different noise levels on FFPMS dataset. As shown in Fig.~\ref{fig:noise}, the proposed S-MIL-T performs well even when there are only $10\%$ fake faces in input instances, which outperforms frame-based XceptionNet and video-based LSTM significantly.
Besides, as shown in Fig.~\ref{fig:gradcam}, the weights get higher on fake faces, which demonstrates the interpretability of the proposed method.

\begin{table}[h]
  \centering
  \small
 \setlength{\tabcolsep}{2mm}{
   \begin{tabular}{ l | c|c|cccc  }
 	\toprule[1.5pt]
    \multirow{2}*{Methods} & \multirow{2}*{DFDC} & \multirow{2}*{Celeb} & \multicolumn{4}{c}{FFPMS} \\ \cline{4-7}
 	~ & ~ & ~ & DF  & F2F & FS & NT \\
 	\hline
 	 	\hline
 	 	 	 XN-avg~\cite{Faceforensics++} & 0.8458 & \textbf{0.9944} & 0.8036 & 0.7714 & 0.8036 & 0.7000 \\
 	 		 XN-max~\cite{Faceforensics++} & 0.7687 & 0.8989  & 0.8536 & 0.7821 & 0.8571 & 0.6571 \\
 	 		 D-FWA~\cite{li2019exposing}  & 0.8511 & 0.9858 & 0.7964 & 0.7571 & 0.8036 & 0.7214 \\
 	 		 LSTM~\cite{hochreiter1997long}   & 0.7902 & 0.9573 & 0.8500 & 0.7564 & 0.7750 & 0.7393 \\
 	 		 I3D~\cite{carreira2017quo}    & 0.8082 & 0.9923 & 0.6214 & 0.6857 & 0.7071 & 0.6679 \\
   \hline
   \hline
               \textbf{S-MIL}        & 0.8378 & \multicolumn{1}{>{\columncolor{mycyan}}l}{0.9923} & \multicolumn{1}{>{\columncolor{mycyan}}l}{0.9036} & \multicolumn{1}{>{\columncolor{mycyan}}l}{0.8107} & \multicolumn{1}{>{\columncolor{mycyan}}l}{0.8609} & \multicolumn{1}{>{\columncolor{pink}}l}{\textbf{0.7857}} \\
             \textbf{S-MIL-T}  & \multicolumn{1}{>{\columncolor{pink}}l}{\textbf{0.8511}} & 0.9884  & \multicolumn{1}{>{\columncolor{pink}}l}{\textbf{0.9071}} & \multicolumn{1}{>{\columncolor{pink}}l}{\textbf{0.8250}} & \multicolumn{1}{>{\columncolor{pink}}l}{\textbf{0.8857}} & \multicolumn{1}{>{\columncolor{mycyan}}l}{0.7535} \\
    \toprule[1.5pt]
   \end{tabular} 
   }
   \caption{Video-level accuracies of the proposed method and other state-of-the-art methods on DFDC, Celeb and FFPMS datasets. Results in bold text indicate the best results while results in pink and blue indicate the best and the second-best results appeared in proposed method.}
  \label{tbl:FFPMS}
  \vspace{-5mm}
\end{table}

\begin{table}[h]
  \centering
  \small
 \setlength{\tabcolsep}{2mm}{
   
\begin{tabular}{c|c|c|cccc}
\toprule[1.5pt]
\multicolumn{2}{c|}{\multirow{2}{*}{Methods}} & \multirow{2}*{CR} & \multicolumn{4}{c}{FaceForensics++} \\ \cline{4-7}
\multicolumn{2}{c|}{~} & ~ & DF & F2F & FS & NT \\ \hline \hline
\multicolumn{7}{c}{Video level accuracy} \\ \hline\hline
Fb & XN-avg~\cite{Faceforensics++} & \multirow{5}*{c0} & 1.0000 & 0.9964 & 0.9964 & 0.9964 \\  \cline{1-2} \cline{4-7} 
\multirow{4}{*}{Vb} & LSTM~\cite{hochreiter1997long} & ~ & 1.0000 & \textbf{1.0000} & 1.0000 & 0.9929 \\ 
~ & I3D~\cite{carreira2017quo} & ~ & 0.9857 & 0.9500 & 0.9714 & 0.9429 \\ 
\cline{2-2} \cline{4-7} 
~ & \textbf{S-MIL} & ~ & 1.0000 & 0.9964 & 1.0000 & 0.9786 \\ 
~ & \textbf{S-MIL-T} & ~ & \multicolumn{1}{>{\columncolor{pink}}l}{\textbf{1.0000}} & \multicolumn{1}{>{\columncolor{mycyan}}l}{0.9964} & \multicolumn{1}{>{\columncolor{pink}}l}{\textbf{1.0000}} & \multicolumn{1}{>{\columncolor{pink}}l}{\textbf{0.9964}} \\  \hline\hline

Fb & XN-avg~\cite{Faceforensics++} & \multirow{5}{*}{c23} & \textbf{1.0000} & 0.9964 & 0.9964 & 0.9393 \\ \cline{1-2} \cline{4-7} 
\multirow{4}{*}{Vb} & LSTM~\cite{hochreiter1997long} & ~ & 0.9964 & 0.9929 & 0.9821 & 0.9393 \\ 
~ & I3D~\cite{carreira2017quo} & ~ & 0.9286 & 0.9286 & 0.9643 & 0.9036 \\ 
\cline{2-2} \cline{4-7} 
~ & \textbf{S-MIL} & ~ & 0.9857 & 0.9929 & 0.9929 & \multicolumn{1}{>{\columncolor{pink}}l}{\textbf{0.9571}} \\ 
~ & \textbf{S-MIL-T} & ~ & \multicolumn{1}{>{\columncolor{mycyan}}l}{0.9964} & \multicolumn{1}{>{\columncolor{pink}}l}{\textbf{0.9964}} & \multicolumn{1}{>{\columncolor{pink}}l}{\textbf{1.0000}} & \multicolumn{1}{>{\columncolor{mycyan}}l}{0.9429} \\  \hline \hline

Fb & XN-avg~\cite{Faceforensics++} & \multirow{5}{*}{c40} & 0.9714 & \textbf{0.9250} & 0.9607 & 0.8607 \\ \cline{1-2} \cline{4-7} 
\multirow{4}{*}{Vb} & LSTM~\cite{hochreiter1997long} & ~ & 0.9643 & 0.8821 & 0.9429 & 0.8821 \\ 
~ & I3D~\cite{carreira2017quo} & ~ & 0.9107 & 0.8643 & 0.9143 & 0.7857 \\ 
\cline{2-2} \cline{4-7} 
~ & \textbf{S-MIL} & ~ & 0.9679 & \multicolumn{1}{>{\columncolor{mycyan}}l}{0.9143} & 0.9464 & \multicolumn{1}{>{\columncolor{pink}}l}{\textbf{0.8857}} \\ 
~ & \textbf{S-MIL-T} & ~ & \multicolumn{1}{>{\columncolor{pink}}l}{\textbf{0.9714}} & 0.9107 & \multicolumn{1}{>{\columncolor{pink}}l}{\textbf{0.9607}} & 0.8679 \\  \hline \hline

\multicolumn{7}{c}{Frame level AUC} \\ \hline\hline
\multirow{2}{*}{Fb} & XN-avg~\cite{Faceforensics++} & \multirow{4}*{c0} & 0.9938 & \textbf{0.9953} & 0.9936 & 0.9729 \\ 
~ & Face X-ray~\cite{X-ray} & ~ & 0.9917 & 0.9906 & 0.9920 & \textbf{0.9893} \\  \cline{1-2} \cline{4-7} 
\multirow{2}{*}{Vb} & MIL~\cite{Revisiting} & ~ & 0.9951 & 0.9859 & 0.9486 & 0.9796 \\  \cline{2-2} \cline{4-7} 
~ & \textbf{S-MIL} & ~ & \multicolumn{1}{>{\columncolor{pink}}l}{\textbf{0.9984}} & \multicolumn{1}{>{\columncolor{mycyan}}l}{0.9934} & \multicolumn{1}{>{\columncolor{pink}}l}{\textbf{0.9961}} & \multicolumn{1}{>{\columncolor{mycyan}}l}{0.9885} \\ 


\toprule[1.5pt]
\end{tabular}
   }
   \caption{Quantitative comparisons with the state-of-the-art methods including frame-based(Fb) and video-based(Vb) algorithms on the FF++ dataset under different compress rates(CR) evaluated by video-level accuracy and frame-level AUC. Results in bold text indicate the best results while results in pink and blue indicate the best and second-best results appeared in proposed method.}
   \label{tbl:all}
 	\vspace{-5mm}
 \end{table}

\begin{table}[h]
  \centering
 \setlength{\tabcolsep}{3mm}{
   
\begin{tabular}{c|c|cccc}
\toprule[1.5pt]
{\multirow{2}{*}{Methods}} & \multirow{2}*{CR} & \multicolumn{4}{c}{FaceForensics++} \\ \cline{3-6}
~ & ~ & DF & F2F & FS & NT \\ \hline \hline
MIL~\cite{Revisiting} & \multirow{3}*{c0} & 1.0000 & 0.9750 & 0.9679 & 0.9357 \\ 
\textbf{S-MIL} & ~ & 1.0000 & 0.9964 & 1.0000 & 0.9786 \\ 
\textbf{S-MIL-T} & ~ & \textbf{1.0000} & \textbf{0.9964} & \textbf{1.0000} & \textbf{0.9964} \\  \hline\hline

MIL~\cite{Revisiting} & \multirow{3}*{c23} & 0.9857 & 0.9857 & 0.9786 & 0.8714 \\ 
\textbf{S-MIL} & ~ & 0.9857 & 0.9929 & 0.9929 & \textbf{0.9571} \\ 
\textbf{S-MIL-T} & ~ & \textbf{0.9964} & \textbf{0.9964} & \textbf{1.0000} & 0.9429 \\  \hline \hline

MIL~\cite{Revisiting} & \multirow{3}*{c40} & 0.9500 & 0.8786 & 0.9071 & 0.8036 \\ 
\textbf{S-MIL} & ~ & 0.9679 & \textbf{0.9143} & 0.9464 & \textbf{0.8857} \\ 
\textbf{S-MIL-T} & ~ & \textbf{0.9714} & 0.9107 & \textbf{0.9607} & 0.8679 \\  

\toprule[1.5pt]
\end{tabular}
   }
   \caption{Ablation study on FF++ dataset under different compress rates evaluated by video-level accuracy in terms of S-MIL and spatial-temporal instances.}
   \label{tbl:ablation}
 	\vspace{-8mm}
 \end{table}

\subsection{Evaluation on Fully Attacked Datasets}

\noindent\textbf{Video-level benchmark results.} 
We test the proposed S-MIL on fully attacked datasets such as Celeb and FF++ to evaluate its generalization ability.
For a comprehensive discussion, we compare the proposed method with a frame-level detector such as XceptionNet~(XN), as well as video-level detectors, {\it i.e.}, LSTM~\cite{hochreiter1997long} and I3D~\cite{carreira2017quo}. As shown in Table~\ref{tbl:FFPMS} and ~\ref{tbl:all}, the proposed multi-instance method outperforms frame-based detectors and video-based detectors in most cases. 

\noindent\textbf{Frame-level benchmark results.} 
Since recent works focus more on frame-level detection, we also evaluate our method to validate the generalization in frame-level detection case.

For this experiment, we take the FaceForensics++~(FF++)\cite{Faceforensics++} dataset as our training data. Only the spatial-level encoding branch is employed for frame-level evaluation since there is no temporal information during testing. Specifically, we sample 20 frames~(less than 10\% of training data) from each video as training data and extract 8 frames from samples randomly to form the video input for S-MIL. During testing, each frame extracted from testing videos is fed into S-MIL individually for frame-level evaluations. We evaluate the resulting four models on this dataset using Area Under Curve~(AUC) metric as frame-level methods~\cite{Faceforensics++, X-ray}.

As shown in Table~\ref{tbl:all}, the proposed S-MIL has achieved state-of-the-art performance with only 10\% training data. This shows that our method can reliably judge input faces.

\begin{figure}[t]
 \centering
  \includegraphics[width=0.8\linewidth]{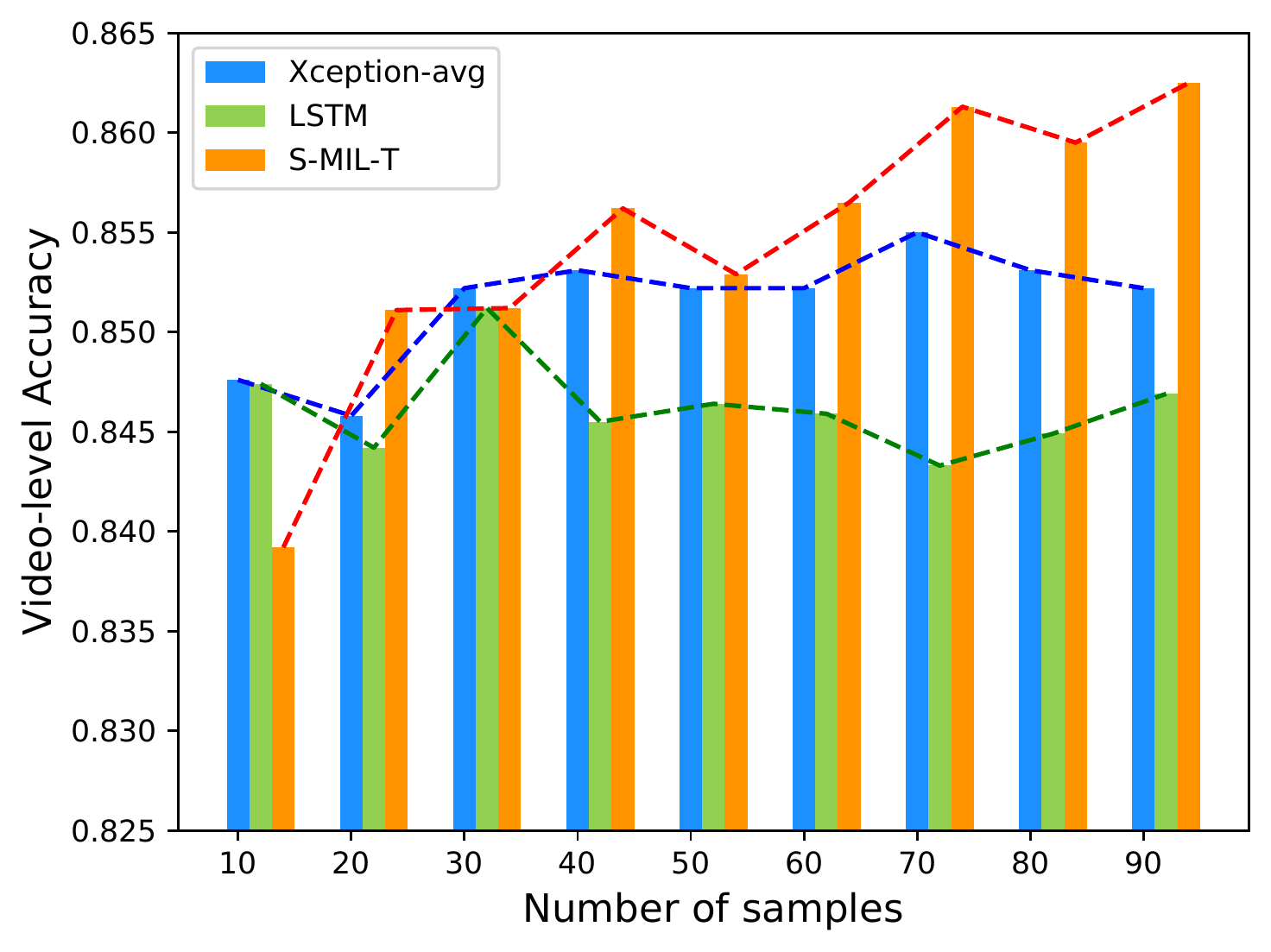}
  \caption{
           Comparison results for DeepFake video detection with different numbers of input instances on DFDC dataset. With more samples, the proposed method can get a better performance while others have little improvements.
}
\label{fig:acc_num}
\end{figure}

\subsection{Ablation Study} 
\noindent\textbf{Component analysis.}
As shown in Table~\ref{tbl:all} and \ref{tbl:ablation}, the proposed S-MIL outperforms the traditional MIL significantly under both frame-level and video-level settings. First, the proposed S-MIL can relieve the gradient vanishing problem, which can help to force every fake instance to be predicted as fake. 
Besides, the proposed weighting mechanism can help to focus on more important instances instead of treating instances uniformly. This can help to gain a reliable prediction. 
Tables~\ref{tbl:FFPMS}, \ref{tbl:all} and \ref{tbl:ablation} show that with the spatial-temporal encoding module, the proposed method can get a better performance, which validates the effectiveness of spatial-temporal instances.

\noindent\textbf{Inference on different number of frames.} 
In video-based settings, for efficiency, sampling is commonly used for less computational and time costs. Thus, the number of sampling frames is an important factor in videos with noises. We explore the influences of different number of frames on frame-based method XceptionNet, video-based method LSTM, and the proposed S-MIL-T with DFDC dataset. 
As shown in Fig.~\ref{fig:acc_num}, the accuracy of proposed S-MIL-T raises when more testing frames are sampled. However, with the growth of sample numbers, the accuracies of XceptionNet and LSTM have little changes, which in turn, shows the superiority of the proposed S-MIL-T.

\section{Conclusion}
In this paper, we introduce a new problem of partially attacked DeepFake video detection which is somewhat ignored in previous study but widely exists in real applications.
We address this problem with MIL paradigm by treating faces and each input video as instances and bag respectively.
Accordingly, we propose S-MIL by building direct connection between bag label prediction and instance embeddings, so that the gradient vanishing problem in traditional MIL can be alleviated. We also conduct theoretic analysis to verify the sharp property of the loss function of S-MIL.
Second, a new spatial-temporal instance is designed to fully model the inconsistency between faces in nearby frames, which helps to improve the detection performance further.
To inspire research on video-level DeepFake detection under partial faces manipulation situations, we construct a new dataset named FFPMS for evaluation of partially attacked DeepFake video detection task.
Experiments on our FFPMS dataset and DFDC benchmark verify that our approach is superior to other counterparts for DeepFake video detection. 
Moreover, our approach can also be adapted to traditional single-frame detection task and achieves state-of-the-art performance on the FF++ datasets, showing a promising result for face information protection.

\newpage

\bibliographystyle{ACM-Reference-Format}
\bibliography{sample-base}


\begin{thebibliography}{56}


\ifx \showCODEN    \undefined \def \showCODEN     #1{\unskip}     \fi
\ifx \showDOI      \undefined \def \showDOI       #1{#1}\fi
\ifx \showISBNx    \undefined \def \showISBNx     #1{\unskip}     \fi
\ifx \showISBNxiii \undefined \def \showISBNxiii  #1{\unskip}     \fi
\ifx \showISSN     \undefined \def \showISSN      #1{\unskip}     \fi
\ifx \showLCCN     \undefined \def \showLCCN      #1{\unskip}     \fi
\ifx \shownote     \undefined \def \shownote      #1{#1}          \fi
\ifx \showarticletitle \undefined \def \showarticletitle #1{#1}   \fi
\ifx \showURL      \undefined \def \showURL       {\relax}        \fi
\providecommand\bibfield[2]{#2}
\providecommand\bibinfo[2]{#2}
\providecommand\natexlab[1]{#1}
\providecommand\showeprint[2][]{arXiv:#2}

\bibitem[\protect\citeauthoryear{Afchar, Nozick, Yamagishi, and Echizen}{Afchar
  et~al\mbox{.}}{2018}]%
        {Mesonet}
\bibfield{author}{\bibinfo{person}{Darius Afchar}, \bibinfo{person}{Vincent
  Nozick}, \bibinfo{person}{Junichi Yamagishi}, {and} \bibinfo{person}{Isao
  Echizen}.} \bibinfo{year}{2018}\natexlab{}.
\newblock \showarticletitle{Mesonet: a compact facial video forgery detection
  network}. In \bibinfo{booktitle}{\emph{IEEE International Workshop on
  Information Forensics and Security (WIFS)}}.
\newblock


\bibitem[\protect\citeauthoryear{Amores}{Amores}{2013}]%
        {Amores2013Multiple}
\bibfield{author}{\bibinfo{person}{Jaume Amores}.}
  \bibinfo{year}{2013}\natexlab{}.
\newblock \showarticletitle{Multiple instance classification: Review, taxonomy
  and comparative study}.
\newblock \bibinfo{journal}{\emph{Artificial Intelligence}}
  \bibinfo{volume}{201} (\bibinfo{year}{2013}), \bibinfo{pages}{81--105}.
\newblock


\bibitem[\protect\citeauthoryear{Anonymous}{Anonymous}{[n.d.]a}]%
        {Partially}
\bibfield{author}{\bibinfo{person}{Anonymous}.}
  \bibinfo{year}{[n.d.]}\natexlab{a}.
\newblock \bibinfo{title}{Example of Partially attacked DeepFake video on
  Youtube}.
\newblock \bibinfo{howpublished}{[EB/OL]}.
\newblock
\newblock
\shownote{\url{https://www.youtube.com/watch?v=BU9YAHigNx8} Accessed April 4,
  2020.}


\bibitem[\protect\citeauthoryear{Anonymous}{Anonymous}{[n.d.]b}]%
        {mtcnn}
\bibfield{author}{\bibinfo{person}{Anonymous}.}
  \bibinfo{year}{[n.d.]}\natexlab{b}.
\newblock \bibinfo{title}{MTCNN in Pytorch}.
\newblock \bibinfo{howpublished}{[EB/OL]}.
\newblock
\newblock
\shownote{\url{https://github.com/timesler/facenet-pytorch} Accessed April 4,
  2020.}


\bibitem[\protect\citeauthoryear{Bahdanau, Cho, and Bengio}{Bahdanau
  et~al\mbox{.}}{2014}]%
        {Bahdanau2014}
\bibfield{author}{\bibinfo{person}{Dzmitry Bahdanau},
  \bibinfo{person}{Kyunghyun Cho}, {and} \bibinfo{person}{Yoshua Bengio}.}
  \bibinfo{year}{2014}\natexlab{}.
\newblock \showarticletitle{Neural machine translation by jointly learning to
  align and translate}. In \bibinfo{booktitle}{\emph{arXiv
  preprint:1409.0473}}.
\newblock


\bibitem[\protect\citeauthoryear{Bayar and Stamm}{Bayar and Stamm}{2016}]%
        {Bayar2016A}
\bibfield{author}{\bibinfo{person}{Belhassen Bayar} {and}
  \bibinfo{person}{Matthew~C Stamm}.} \bibinfo{year}{2016}\natexlab{}.
\newblock \showarticletitle{A Deep Learning Approach to Universal Image
  Manipulation Detection Using a New Convolutional Layer}. In
  \bibinfo{booktitle}{\emph{the 4th ACM Workshop}}. \bibinfo{pages}{5--10}.
\newblock


\bibitem[\protect\citeauthoryear{Carreira and Zisserman}{Carreira and
  Zisserman}{2017}]%
        {carreira2017quo}
\bibfield{author}{\bibinfo{person}{Joao Carreira} {and} \bibinfo{person}{Andrew
  Zisserman}.} \bibinfo{year}{2017}\natexlab{}.
\newblock \showarticletitle{Quo vadis, action recognition? a new model and the
  kinetics dataset}. In \bibinfo{booktitle}{\emph{CVPR}}.
  \bibinfo{pages}{6299--6308}.
\newblock


\bibitem[\protect\citeauthoryear{Chen and Guestrin}{Chen and Guestrin}{2016}]%
        {chen2016xgboost}
\bibfield{author}{\bibinfo{person}{Tianqi Chen} {and} \bibinfo{person}{Carlos
  Guestrin}.} \bibinfo{year}{2016}\natexlab{}.
\newblock \showarticletitle{Xgboost: A scalable tree boosting system}. In
  \bibinfo{booktitle}{\emph{KDD}}. \bibinfo{pages}{785--794}.
\newblock


\bibitem[\protect\citeauthoryear{Cozzolino, Poggi, and Verdoliva}{Cozzolino
  et~al\mbox{.}}{2017}]%
        {CozzolinoRecasting}
\bibfield{author}{\bibinfo{person}{Davide Cozzolino}, \bibinfo{person}{Giovanni
  Poggi}, {and} \bibinfo{person}{Luisa Verdoliva}.}
  \bibinfo{year}{2017}\natexlab{}.
\newblock \showarticletitle{Recasting Residual-based Local Descriptors as
  Convolutional Neural Networks: an Application to Image Forgery Detection}.
\newblock  (\bibinfo{year}{2017}), \bibinfo{pages}{159--164}.
\newblock


\bibitem[\protect\citeauthoryear{DeepFakes}{DeepFakes}{2019a}]%
        {DeepFakes}
\bibfield{author}{\bibinfo{person}{DeepFakes}.}
  \bibinfo{year}{2019}\natexlab{a}.
\newblock \showarticletitle{www.github.com/deepfakes/ faceswap}.
\newblock \bibinfo{journal}{\emph{Accessed}} (\bibinfo{year}{2019}).
\newblock


\bibitem[\protect\citeauthoryear{DeepFakes}{DeepFakes}{2019b}]%
        {FaceSwap}
\bibfield{author}{\bibinfo{person}{DeepFakes}.}
  \bibinfo{year}{2019}\natexlab{b}.
\newblock \showarticletitle{www.github.com/MarekKowalski/}.
\newblock \bibinfo{journal}{\emph{Accessed}} (\bibinfo{year}{2019}).
\newblock


\bibitem[\protect\citeauthoryear{Dietterich, Lathrop, and
  Lozano-Pérez}{Dietterich et~al\mbox{.}}{2001}]%
        {Dietterich2001Solving}
\bibfield{author}{\bibinfo{person}{Thomas~G. Dietterich},
  \bibinfo{person}{Richard~H. Lathrop}, {and} \bibinfo{person}{Tomás
  Lozano-Pérez}.} \bibinfo{year}{2001}\natexlab{}.
\newblock \showarticletitle{Solving the multiple-instance problem with
  axis-parallel rectangles}.
\newblock \bibinfo{journal}{\emph{Artificial Intelligence}}
  \bibinfo{volume}{89}, \bibinfo{number}{1-2} (\bibinfo{year}{2001}),
  \bibinfo{pages}{31--71}.
\newblock


\bibitem[\protect\citeauthoryear{Ding, Raziei, Larson, Olinick, Krueger, and
  Hahsler}{Ding et~al\mbox{.}}{2019}]%
        {DingSwapped}
\bibfield{author}{\bibinfo{person}{Xinyi Ding}, \bibinfo{person}{Zohreh
  Raziei}, \bibinfo{person}{Eric~C Larson}, \bibinfo{person}{Eli~V Olinick},
  \bibinfo{person}{Paul~S Krueger}, {and} \bibinfo{person}{Michael Hahsler}.}
  \bibinfo{year}{2019}\natexlab{}.
\newblock \showarticletitle{Swapped Face Detection using Deep Learning and
  Subjective Assessment.}
\newblock \bibinfo{journal}{\emph{arXiv: Learning}} (\bibinfo{year}{2019}).
\newblock


\bibitem[\protect\citeauthoryear{Dolhansky, Howes, Pflaum, Baram, and
  Ferrer}{Dolhansky et~al\mbox{.}}{2019}]%
        {dolhansky2019deepfake}
\bibfield{author}{\bibinfo{person}{Brian Dolhansky}, \bibinfo{person}{Russ
  Howes}, \bibinfo{person}{Ben Pflaum}, \bibinfo{person}{Nicole Baram}, {and}
  \bibinfo{person}{Cristian~Canton Ferrer}.} \bibinfo{year}{2019}\natexlab{}.
\newblock \showarticletitle{The Deepfake Detection Challenge (DFDC) Preview
  Dataset}.
\newblock \bibinfo{journal}{\emph{arXiv preprint:1910.08854}}
  (\bibinfo{year}{2019}).
\newblock


\bibitem[\protect\citeauthoryear{Facebook}{Facebook}{[n.d.]}]%
        {DFDChallenge}
\bibfield{author}{\bibinfo{person}{Facebook}.}
  \bibinfo{year}{[n.d.]}\natexlab{}.
\newblock \bibinfo{title}{DeepFake Detection Challenge}.
\newblock \bibinfo{howpublished}{[EB/OL]}.
\newblock
\newblock
\shownote{\url{https://www.kaggle.com/c/deepfake-detection-challenge} Accessed
  May 20, 2020.}


\bibitem[\protect\citeauthoryear{Feng, Ni, Tian, and Yan}{Feng
  et~al\mbox{.}}{2011}]%
        {Feng2011Geometric}
\bibfield{author}{\bibinfo{person}{Jiashi Feng}, \bibinfo{person}{Bingbing Ni},
  \bibinfo{person}{Qi Tian}, {and} \bibinfo{person}{Shuicheng Yan}.}
  \bibinfo{year}{2011}\natexlab{}.
\newblock \showarticletitle{Geometric lp-norm feature pooling for image
  classification}. In \bibinfo{booktitle}{\emph{CVPR}}.
\newblock


\bibitem[\protect\citeauthoryear{Feng and Zhou}{Feng and Zhou}{2017}]%
        {MIML}
\bibfield{author}{\bibinfo{person}{Ji Feng} {and} \bibinfo{person}{Zhihua
  Zhou}.} \bibinfo{year}{2017}\natexlab{}.
\newblock \showarticletitle{Deep MIML Network}. In
  \bibinfo{booktitle}{\emph{AAAI}}.
\newblock


\bibitem[\protect\citeauthoryear{Fridrich and Kodovsky}{Fridrich and
  Kodovsky}{2012}]%
        {Rich}
\bibfield{author}{\bibinfo{person}{Jessica Fridrich} {and} \bibinfo{person}{Jan
  Kodovsky}.} \bibinfo{year}{2012}\natexlab{}.
\newblock \showarticletitle{Rich Models for Steganalysis of Digital Images}.
\newblock \bibinfo{journal}{\emph{IEEE Transactions on Information Forensics
  and Security}} \bibinfo{volume}{7}, \bibinfo{number}{3},
  \bibinfo{pages}{868--882}.
\newblock


\bibitem[\protect\citeauthoryear{Goljan and Fridrich}{Goljan and
  Fridrich}{2015}]%
        {CFA-aware}
\bibfield{author}{\bibinfo{person}{Miroslav Goljan} {and}
  \bibinfo{person}{Jessica Fridrich}.} \bibinfo{year}{2015}\natexlab{}.
\newblock \showarticletitle{CFA-aware features for steganalysis of color
  images}.
\newblock \bibinfo{journal}{\emph{electronic imaging}}  \bibinfo{volume}{9409}.
\newblock


\bibitem[\protect\citeauthoryear{Goodfellow, Pouget-Abadie, Mirza, Xu,
  Warde-Farley, Ozair, Courville, and Bengio}{Goodfellow et~al\mbox{.}}{2014}]%
        {goodfellow2014generative}
\bibfield{author}{\bibinfo{person}{Ian Goodfellow}, \bibinfo{person}{Jean
  Pouget-Abadie}, \bibinfo{person}{Mehdi Mirza}, \bibinfo{person}{Bing Xu},
  \bibinfo{person}{David Warde-Farley}, \bibinfo{person}{Sherjil Ozair},
  \bibinfo{person}{Aaron Courville}, {and} \bibinfo{person}{Yoshua Bengio}.}
  \bibinfo{year}{2014}\natexlab{}.
\newblock \showarticletitle{Generative adversarial nets}. In
  \bibinfo{booktitle}{\emph{NIPS}}. \bibinfo{pages}{2672--2680}.
\newblock


\bibitem[\protect\citeauthoryear{Guera and Delp}{Guera and Delp}{2018}]%
        {Guera}
\bibfield{author}{\bibinfo{person}{David Guera} {and} \bibinfo{person}{Edward~J
  Delp}.} \bibinfo{year}{2018}\natexlab{}.
\newblock \showarticletitle{Deepfake Video Detection Using Recurrent Neural
  Networks}. In \bibinfo{booktitle}{\emph{AVSS}}. \bibinfo{pages}{1--6}.
\newblock


\bibitem[\protect\citeauthoryear{Hochreiter and Schmidhuber}{Hochreiter and
  Schmidhuber}{1997}]%
        {hochreiter1997long}
\bibfield{author}{\bibinfo{person}{Sepp Hochreiter} {and}
  \bibinfo{person}{J{\"u}rgen Schmidhuber}.} \bibinfo{year}{1997}\natexlab{}.
\newblock \showarticletitle{Long short-term memory}.
\newblock \bibinfo{journal}{\emph{Neural computation}} \bibinfo{volume}{9},
  \bibinfo{number}{8} (\bibinfo{year}{1997}), \bibinfo{pages}{1735--1780}.
\newblock


\bibitem[\protect\citeauthoryear{Ilse, Tomczak, and Welling}{Ilse
  et~al\mbox{.}}{2018}]%
        {ilse2018attention}
\bibfield{author}{\bibinfo{person}{Maximilian Ilse}, \bibinfo{person}{Jakub~M
  Tomczak}, {and} \bibinfo{person}{Max Welling}.}
  \bibinfo{year}{2018}\natexlab{}.
\newblock \showarticletitle{Attention-based deep multiple instance learning}.
\newblock \bibinfo{journal}{\emph{arXiv preprint:1802.04712}}
  (\bibinfo{year}{2018}).
\newblock


\bibitem[\protect\citeauthoryear{Keeler, Rumelhart, and Leow}{Keeler
  et~al\mbox{.}}{1990}]%
        {Keeler1990Integrated}
\bibfield{author}{\bibinfo{person}{James~D. Keeler}, \bibinfo{person}{David~E.
  Rumelhart}, {and} \bibinfo{person}{Wee~Kheng Leow}.}
  \bibinfo{year}{1990}\natexlab{}.
\newblock \showarticletitle{Integrated segmentation and recognition of
  hand-printed numerals}. In \bibinfo{booktitle}{\emph{NIPS}}.
\newblock


\bibitem[\protect\citeauthoryear{Kim}{Kim}{2014}]%
        {kim2014convolutional}
\bibfield{author}{\bibinfo{person}{Yoon Kim}.} \bibinfo{year}{2014}\natexlab{}.
\newblock \showarticletitle{Convolutional neural networks for sentence
  classification}.
\newblock \bibinfo{journal}{\emph{arXiv preprint:1408.5882}}
  (\bibinfo{year}{2014}).
\newblock


\bibitem[\protect\citeauthoryear{Kingma and Ba}{Kingma and Ba}{2014}]%
        {kingma2014adam}
\bibfield{author}{\bibinfo{person}{Diederik~P Kingma} {and}
  \bibinfo{person}{Jimmy Ba}.} \bibinfo{year}{2014}\natexlab{}.
\newblock \showarticletitle{Adam: A method for stochastic optimization}.
\newblock \bibinfo{journal}{\emph{arXiv preprint:1412.6980}}
  (\bibinfo{year}{2014}).
\newblock


\bibitem[\protect\citeauthoryear{Li, Bao, Zhang, Yang, Chen, Wen, and Guo}{Li
  et~al\mbox{.}}{2019a}]%
        {X-ray}
\bibfield{author}{\bibinfo{person}{Lingzhi Li}, \bibinfo{person}{Jianmin Bao},
  \bibinfo{person}{Ting Zhang}, \bibinfo{person}{Hao Yang},
  \bibinfo{person}{Dong Chen}, \bibinfo{person}{Fang Wen}, {and}
  \bibinfo{person}{Baining Guo}.} \bibinfo{year}{2019}\natexlab{a}.
\newblock \showarticletitle{Face X-ray for More General Face Forgery
  Detection}.
\newblock \bibinfo{journal}{\emph{CVPR}}.
\newblock


\bibitem[\protect\citeauthoryear{Li and Lyu}{Li and Lyu}{2019}]%
        {li2019exposing}
\bibfield{author}{\bibinfo{person}{Yuezun Li} {and} \bibinfo{person}{Siwei
  Lyu}.} \bibinfo{year}{2019}\natexlab{}.
\newblock \showarticletitle{Exposing DeepFake Videos By Detecting Face Warping
  Artifacts}. In \bibinfo{booktitle}{\emph{CVPRW}}.
\newblock


\bibitem[\protect\citeauthoryear{Li, Yang, Sun, Qi, and Lyu}{Li
  et~al\mbox{.}}{2019b}]%
        {li2019celeb}
\bibfield{author}{\bibinfo{person}{Yuezun Li}, \bibinfo{person}{Xin Yang},
  \bibinfo{person}{Pu Sun}, \bibinfo{person}{Honggang Qi}, {and}
  \bibinfo{person}{Siwei Lyu}.} \bibinfo{year}{2019}\natexlab{b}.
\newblock \showarticletitle{Celeb-df: A new dataset for deepfake forensics}.
\newblock \bibinfo{journal}{\emph{arXiv preprint:1909.12962}}
  (\bibinfo{year}{2019}).
\newblock


\bibitem[\protect\citeauthoryear{Lin, Goyal, Girshick, He, and Doll{\'a}r}{Lin
  et~al\mbox{.}}{2017b}]%
        {lin2017focal}
\bibfield{author}{\bibinfo{person}{Tsung-Yi Lin}, \bibinfo{person}{Priya
  Goyal}, \bibinfo{person}{Ross Girshick}, \bibinfo{person}{Kaiming He}, {and}
  \bibinfo{person}{Piotr Doll{\'a}r}.} \bibinfo{year}{2017}\natexlab{b}.
\newblock \showarticletitle{Focal loss for dense object detection}. In
  \bibinfo{booktitle}{\emph{ICCV}}. \bibinfo{pages}{2980--2988}.
\newblock


\bibitem[\protect\citeauthoryear{Lin, Feng, Santos, Yu, Xiang, Zhou, and
  Bengio}{Lin et~al\mbox{.}}{2017a}]%
        {Lin2017}
\bibfield{author}{\bibinfo{person}{Zhouhan Lin}, \bibinfo{person}{Minwei Feng},
  \bibinfo{person}{Cicero Nogueira~Dos Santos}, \bibinfo{person}{Mo Yu},
  \bibinfo{person}{Bing Xiang}, \bibinfo{person}{Bowen Zhou}, {and}
  \bibinfo{person}{Yoshua Bengio}.} \bibinfo{year}{2017}\natexlab{a}.
\newblock \showarticletitle{A Structured Self-attentive Sentence Embedding}.
\newblock \bibinfo{journal}{\emph{arXiv: Computation and Language}}.
\newblock


\bibitem[\protect\citeauthoryear{Maron and Lozano-Perez}{Maron and
  Lozano-Perez}{1998}]%
        {Maron1998}
\bibfield{author}{\bibinfo{person}{Oded Maron} {and} \bibinfo{person}{Tomas
  Lozano-Perez}.} \bibinfo{year}{1998}\natexlab{}.
\newblock \showarticletitle{A framework for multiple-instance learning}. In
  \bibinfo{booktitle}{\emph{NIPS}}.
\newblock


\bibitem[\protect\citeauthoryear{Marra, Gragnaniello, Cozzolino, and
  Verdoliva}{Marra et~al\mbox{.}}{2018}]%
        {Marra2018Detection}
\bibfield{author}{\bibinfo{person}{Francesco Marra}, \bibinfo{person}{Diego
  Gragnaniello}, \bibinfo{person}{Davide Cozzolino}, {and}
  \bibinfo{person}{Luisa Verdoliva}.} \bibinfo{year}{2018}\natexlab{}.
\newblock \showarticletitle{Detection of GAN-Generated Fake Images over Social
  Networks}. In \bibinfo{booktitle}{\emph{IEEE MIPR}}.
\newblock


\bibitem[\protect\citeauthoryear{Oquab, Bottou, Laptev, and Sivic}{Oquab
  et~al\mbox{.}}{2014}]%
        {Oquab}
\bibfield{author}{\bibinfo{person}{Maxime Oquab}, \bibinfo{person}{Leon
  Bottou}, \bibinfo{person}{Ivan Laptev}, {and} \bibinfo{person}{Josef Sivic}.}
  \bibinfo{year}{2014}\natexlab{}.
\newblock \showarticletitle{Weakly supervised object recognition with
  convolutional neural networks}. In \bibinfo{booktitle}{\emph{NIPS}}.
\newblock


\bibitem[\protect\citeauthoryear{Pan, Zhang, and Lyu}{Pan
  et~al\mbox{.}}{2012}]%
        {Exposing}
\bibfield{author}{\bibinfo{person}{Xunyu Pan}, \bibinfo{person}{Xing Zhang},
  {and} \bibinfo{person}{Siwei Lyu}.} \bibinfo{year}{2012}\natexlab{}.
\newblock \showarticletitle{Exposing image splicing with inconsistent local
  noise variances}. In \bibinfo{booktitle}{\emph{ICCP}}.
  \bibinfo{pages}{1--10}.
\newblock


\bibitem[\protect\citeauthoryear{Pappas and Popescu-Belis}{Pappas and
  Popescu-Belis}{2014}]%
        {Explaining}
\bibfield{author}{\bibinfo{person}{Nikolaos Pappas} {and}
  \bibinfo{person}{Andrei Popescu-Belis}.} \bibinfo{year}{2014}\natexlab{}.
\newblock \showarticletitle{Explaining the stars: Weighted multiple-instance
  learning for aspectbased sentiment analysis}. In
  \bibinfo{booktitle}{\emph{EMNLP}}.
\newblock


\bibitem[\protect\citeauthoryear{Pappas and Popescu-Belis}{Pappas and
  Popescu-Belis}{2017}]%
        {Explicit}
\bibfield{author}{\bibinfo{person}{Nikolaos Pappas} {and}
  \bibinfo{person}{Andrei Popescu-Belis}.} \bibinfo{year}{2017}\natexlab{}.
\newblock \showarticletitle{Explicit Document Modeling through Weighted
  Multiple-Instance Learning}. In \bibinfo{booktitle}{\emph{Journal of
  Artificial Intelligence Research}}.
\newblock


\bibitem[\protect\citeauthoryear{Pinheiro and Collobert}{Pinheiro and
  Collobert}{2015}]%
        {image-level}
\bibfield{author}{\bibinfo{person}{Pedro~O Pinheiro} {and}
  \bibinfo{person}{Ronan Collobert}.} \bibinfo{year}{2015}\natexlab{}.
\newblock \showarticletitle{From image-level to pixel-level labeling with
  convolutional networks}. In \bibinfo{booktitle}{\emph{CVPR}}.
\newblock


\bibitem[\protect\citeauthoryear{Quellec, Cazuguel, Cochener, and
  Lamard}{Quellec et~al\mbox{.}}{2017}]%
        {medical}
\bibfield{author}{\bibinfo{person}{Gwenole Quellec}, \bibinfo{person}{Guy
  Cazuguel}, \bibinfo{person}{Beatrice Cochener}, {and}
  \bibinfo{person}{Mathieu Lamard}.} \bibinfo{year}{2017}\natexlab{}.
\newblock \showarticletitle{Multiple-Instance Learning for Medical Image and
  Video Analysis}.
\newblock \bibinfo{journal}{\emph{IEEE Reviews in Biomedical Engineering}}
  \bibinfo{volume}{10}, \bibinfo{pages}{213--234}.
\newblock


\bibitem[\protect\citeauthoryear{Radford, Metz, and Chintala}{Radford
  et~al\mbox{.}}{2015}]%
        {radford2015unsupervised}
\bibfield{author}{\bibinfo{person}{Alec Radford}, \bibinfo{person}{Luke Metz},
  {and} \bibinfo{person}{Soumith Chintala}.} \bibinfo{year}{2015}\natexlab{}.
\newblock \showarticletitle{Unsupervised representation learning with deep
  convolutional generative adversarial networks}.
\newblock \bibinfo{journal}{\emph{arXiv preprint:1511.06434}}
  (\bibinfo{year}{2015}).
\newblock


\bibitem[\protect\citeauthoryear{Rahmouni, Nozick, Yamagishi, and
  Echizen}{Rahmouni et~al\mbox{.}}{2017}]%
        {Rahmouni2017Distinguishing}
\bibfield{author}{\bibinfo{person}{Nicolas Rahmouni}, \bibinfo{person}{Vincent
  Nozick}, \bibinfo{person}{Junichi Yamagishi}, {and} \bibinfo{person}{Isao
  Echizen}.} \bibinfo{year}{2017}\natexlab{}.
\newblock \showarticletitle{Distinguishing computer graphics from natural
  images using convolution neural networks}. In \bibinfo{booktitle}{\emph{IEEE
  Workshop on Information Forensics and Security}}.
\newblock


\bibitem[\protect\citeauthoryear{Rossler, Cozzolino, Verdoliva, Riess, Thies,
  and Niesner}{Rossler et~al\mbox{.}}{2018}]%
        {Faceforensics}
\bibfield{author}{\bibinfo{person}{Andreas Rossler}, \bibinfo{person}{Davide
  Cozzolino}, \bibinfo{person}{Luisa Verdoliva}, \bibinfo{person}{Christian
  Riess}, \bibinfo{person}{Justus Thies}, {and} \bibinfo{person}{Matthias
  Niesner}.} \bibinfo{year}{2018}\natexlab{}.
\newblock \showarticletitle{FaceForensics: A Large-scale Video Dataset for
  Forgery Detection in Human Faces}.
\newblock \bibinfo{journal}{\emph{CVPR}}.
\newblock


\bibitem[\protect\citeauthoryear{Rossler, Cozzolino, Verdoliva, Riess, Thies,
  and Niesner}{Rossler et~al\mbox{.}}{2019}]%
        {Faceforensics++}
\bibfield{author}{\bibinfo{person}{Andreas Rossler}, \bibinfo{person}{Davide
  Cozzolino}, \bibinfo{person}{Luisa Verdoliva}, \bibinfo{person}{Christian
  Riess}, \bibinfo{person}{Justus Thies}, {and} \bibinfo{person}{Matthias
  Niesner}.} \bibinfo{year}{2019}\natexlab{}.
\newblock \showarticletitle{Faceforensics++: Learning to detect manipulated
  facial images}. In \bibinfo{booktitle}{\emph{arXiv preprint
  arXiv:1901.08971}}.
\newblock


\bibitem[\protect\citeauthoryear{Rössler}{Rössler}{[n.d.]}]%
        {dlib}
\bibfield{author}{\bibinfo{person}{Andreas Rössler}.}
  \bibinfo{year}{[n.d.]}\natexlab{}.
\newblock \bibinfo{title}{Face Detector used in Xception on Github}.
\newblock \bibinfo{howpublished}{[EB/OL]}.
\newblock
\newblock
\shownote{\url{https://github.com/ondyari/FaceForensics} Accessed April 4,
  2020.}


\bibitem[\protect\citeauthoryear{Sabir, Cheng, Jaiswal, AbdAlmageed, Masi, and
  Natarajan}{Sabir et~al\mbox{.}}{2019}]%
        {sabir2019recurrent}
\bibfield{author}{\bibinfo{person}{Ekraam Sabir}, \bibinfo{person}{Jiaxin
  Cheng}, \bibinfo{person}{Ayush Jaiswal}, \bibinfo{person}{Wael AbdAlmageed},
  \bibinfo{person}{Iacopo Masi}, {and} \bibinfo{person}{Prem Natarajan}.}
  \bibinfo{year}{2019}\natexlab{}.
\newblock \showarticletitle{Recurrent convolutional strategies for face
  manipulation detection in videos}.
\newblock \bibinfo{journal}{\emph{Interfaces (GUI)}}  \bibinfo{volume}{3}
  (\bibinfo{year}{2019}), \bibinfo{pages}{1}.
\newblock


\bibitem[\protect\citeauthoryear{Selvaraju, Cogswell, Das, Vedantam, Parikh,
  and Batra}{Selvaraju et~al\mbox{.}}{2017}]%
        {selvaraju2017grad}
\bibfield{author}{\bibinfo{person}{Ramprasaath~R Selvaraju},
  \bibinfo{person}{Michael Cogswell}, \bibinfo{person}{Abhishek Das},
  \bibinfo{person}{Ramakrishna Vedantam}, \bibinfo{person}{Devi Parikh}, {and}
  \bibinfo{person}{Dhruv Batra}.} \bibinfo{year}{2017}\natexlab{}.
\newblock \showarticletitle{Grad-cam: Visual explanations from deep networks
  via gradient-based localization}. In \bibinfo{booktitle}{\emph{ICCV}}.
  \bibinfo{pages}{618--626}.
\newblock


\bibitem[\protect\citeauthoryear{Sirinukunwattana, Raza, Tsang, Snead, Cree,
  and Rajpoot}{Sirinukunwattana et~al\mbox{.}}{2016}]%
        {Sirinukunwattana2016Locality}
\bibfield{author}{\bibinfo{person}{Korsuk Sirinukunwattana},
  \bibinfo{person}{Shan Raza}, \bibinfo{person}{Yee~Wah Tsang},
  \bibinfo{person}{David Snead}, \bibinfo{person}{Ian Cree}, {and}
  \bibinfo{person}{Nasir Rajpoot}.} \bibinfo{year}{2016}\natexlab{}.
\newblock \showarticletitle{Locality sensitive deep learning for detection and
  classification of nuclei in routine colon cancer histology images}.
\newblock \bibinfo{journal}{\emph{IEEE Transactions on Medical Imaging}}
  \bibinfo{volume}{35}, \bibinfo{number}{5} (\bibinfo{year}{2016}),
  \bibinfo{pages}{1196--1206}.
\newblock


\bibitem[\protect\citeauthoryear{Tariq, Lee, Kim, Shin, and Woo}{Tariq
  et~al\mbox{.}}{2018}]%
        {Shahroz}
\bibfield{author}{\bibinfo{person}{Shahroz Tariq}, \bibinfo{person}{Sangyup
  Lee}, \bibinfo{person}{Hoyoung Kim}, \bibinfo{person}{Youjin Shin}, {and}
  \bibinfo{person}{Simon~S Woo}.} \bibinfo{year}{2018}\natexlab{}.
\newblock \showarticletitle{Detecting both machine and human created fake face
  images in the wild}.
\newblock \bibinfo{journal}{\emph{In Proceedings of the 2nd International
  Workshop on Multimedia Privacy and Security}} (\bibinfo{year}{2018}).
\newblock


\bibitem[\protect\citeauthoryear{Thies, Zollhofer, and Niesner}{Thies
  et~al\mbox{.}}{2019}]%
        {Deferred}
\bibfield{author}{\bibinfo{person}{Justus Thies}, \bibinfo{person}{Michael
  Zollhofer}, {and} \bibinfo{person}{Matthias Niesner}.}
  \bibinfo{year}{2019}\natexlab{}.
\newblock \showarticletitle{Deferred neural rendering: image synthesis using
  neural textures}.
\newblock \bibinfo{journal}{\emph{ACM Transactions on Graphics}}
  \bibinfo{volume}{38}, \bibinfo{number}{4} (\bibinfo{year}{2019}),
  \bibinfo{pages}{66}.
\newblock


\bibitem[\protect\citeauthoryear{Thies, Zollhofer, Stamminger, Theobalt, and
  Niesner}{Thies et~al\mbox{.}}{2018}]%
        {Justus2018Face2Face}
\bibfield{author}{\bibinfo{person}{Justus Thies}, \bibinfo{person}{Michael
  Zollhofer}, \bibinfo{person}{Marc Stamminger}, \bibinfo{person}{Christian
  Theobalt}, {and} \bibinfo{person}{Matthias Niesner}.}
  \bibinfo{year}{2018}\natexlab{}.
\newblock \showarticletitle{Face2Face: real-time face capture and reenactment
  of RGB videos}.
\newblock \bibinfo{journal}{\emph{Communications of The ACM}}
  \bibinfo{volume}{62}, \bibinfo{number}{1} (\bibinfo{year}{2018}),
  \bibinfo{pages}{96--104}.
\newblock


\bibitem[\protect\citeauthoryear{Wang, Ma, Juefeixu, Xie, Wang, and Liu}{Wang
  et~al\mbox{.}}{2019}]%
        {Fakespotter}
\bibfield{author}{\bibinfo{person}{Run Wang}, \bibinfo{person}{Lei Ma},
  \bibinfo{person}{Felix Juefeixu}, \bibinfo{person}{Xiaofei Xie},
  \bibinfo{person}{Jian Wang}, {and} \bibinfo{person}{Yang Liu}.}
  \bibinfo{year}{2019}\natexlab{}.
\newblock \showarticletitle{Fakespotter: A simple baseline for spotting
  ai-synthesized fake faces}. In \bibinfo{booktitle}{\emph{arXiv preprint
  arXiv:1909.06122}}.
\newblock


\bibitem[\protect\citeauthoryear{Wang, Yan, Tang, Bai, and Liu}{Wang
  et~al\mbox{.}}{2016}]%
        {Revisiting}
\bibfield{author}{\bibinfo{person}{Xinggang Wang}, \bibinfo{person}{Yongluan
  Yan}, \bibinfo{person}{Peng Tang}, \bibinfo{person}{Xiang Bai}, {and}
  \bibinfo{person}{Wenyu Liu}.} \bibinfo{year}{2016}\natexlab{}.
\newblock \showarticletitle{Revisiting multiple instance neural networks}. In
  \bibinfo{booktitle}{\emph{Pattern Recognition}}.
\newblock


\bibitem[\protect\citeauthoryear{Wang, Yan, Tang, Bai, and Liu}{Wang
  et~al\mbox{.}}{2018}]%
        {wang2018revisiting}
\bibfield{author}{\bibinfo{person}{Xinggang Wang}, \bibinfo{person}{Yongluan
  Yan}, \bibinfo{person}{Peng Tang}, \bibinfo{person}{Xiang Bai}, {and}
  \bibinfo{person}{Wenyu Liu}.} \bibinfo{year}{2018}\natexlab{}.
\newblock \showarticletitle{Revisiting multiple instance neural networks}.
\newblock \bibinfo{journal}{\emph{Pattern Recognition}}  \bibinfo{volume}{74}
  (\bibinfo{year}{2018}), \bibinfo{pages}{15--24}.
\newblock


\bibitem[\protect\citeauthoryear{Xu, Ba, Kiros, Cho, Courville, Salakhudinov,
  Zemel, and Bengio}{Xu et~al\mbox{.}}{2015}]%
        {Show2015}
\bibfield{author}{\bibinfo{person}{Kelvin Xu}, \bibinfo{person}{Jimmy Ba},
  \bibinfo{person}{Ryan Kiros}, \bibinfo{person}{Kyunghyun Cho},
  \bibinfo{person}{Aaron Courville}, \bibinfo{person}{Ruslan Salakhudinov},
  \bibinfo{person}{Rich Zemel}, {and} \bibinfo{person}{Yoshua Bengio}.}
  \bibinfo{year}{2015}\natexlab{}.
\newblock \showarticletitle{Show, attend and tell: Neural image caption
  generation with visual attention}. In \bibinfo{booktitle}{\emph{ICML}}.
\newblock


\bibitem[\protect\citeauthoryear{Zagoruyko and Komodakis}{Zagoruyko and
  Komodakis}{2016}]%
        {ZagoruykoPaying}
\bibfield{author}{\bibinfo{person}{Sergey Zagoruyko} {and}
  \bibinfo{person}{Nikos Komodakis}.} \bibinfo{year}{2016}\natexlab{}.
\newblock \showarticletitle{Paying more attention to attention: Improving the
  performance of convolutional neural networks via attention transfer}. In
  \bibinfo{booktitle}{\emph{arXiv preprint}}.
\newblock


\bibitem[\protect\citeauthoryear{Zhu, Lou, Vang, and Xie}{Zhu
  et~al\mbox{.}}{2017}]%
        {Deepzhu}
\bibfield{author}{\bibinfo{person}{Wentao Zhu}, \bibinfo{person}{Qi Lou},
  \bibinfo{person}{Yeeleng~Scott Vang}, {and} \bibinfo{person}{Xiaohui Xie}.}
  \bibinfo{year}{2017}\natexlab{}.
\newblock \showarticletitle{Deep multi-instance networks with sparse label
  assignment for whole mammogram classification}. In
  \bibinfo{booktitle}{\emph{MICCAI}}.
\newblock


\end{thebibliography}

\newpage
\section{Appendix}

\begin{figure*}[!t]
 \centering
  \includegraphics[width=1.0\linewidth]{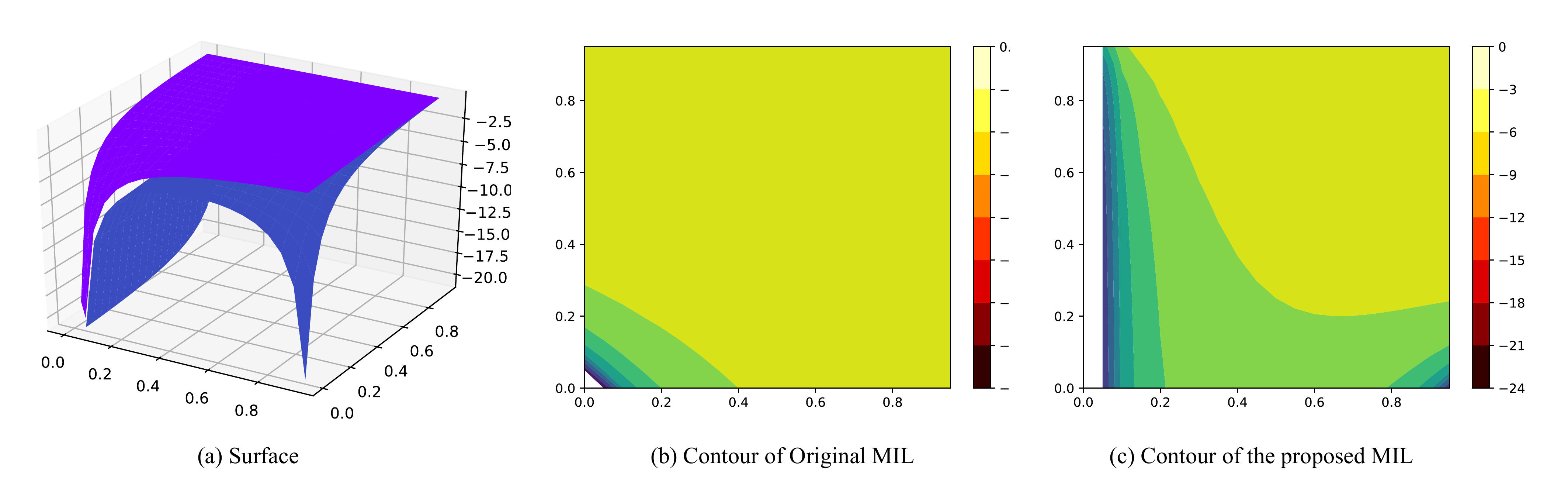}
  \caption{
          Visualization for MIL and the proposed MIL.
}
\label{fig:contour}
\end{figure*}


\subsection{Proof}

Section 3.2 has shown an alternative formulation for Multi Instance Learning (MIL), which is given by:
\begin{equation}
\label{eq:1}
p = \frac{1}{1+\prod_{j=1}^M(\frac{1}{p^{j}}-1)},
\end{equation}
where $p^{j}$ is the output probability of $j$-th instance and $M$ is the total number of instances. Formula~\ref{eq:1} merges the prediction of all instances and outputs the final probability $p$. The final training objective related to $p$ is the cross entropy loss:
\begin{equation}
\label{eq:2}
\mathcal{L}(p) = -y\log(p)-(1-y)\log(1-p),
\end{equation}
where $y$ is the ground-truth. For the convenience of derivation, we present the traditional MIL method as follows:
\begin{equation}
\label{eq:3}
\hat{p} = 1-\prod_{j=1}^M(1-p^{j}).
\end{equation}
In order to make a better distinction, we set $\hat{p}$ as the output of traditional MIL. Note that traditional MIL also has the same objective function as Eq.~\ref{eq:2}.

Next we explain why the newly proposed MIL paradigm could relieve the gradient vanishing in positive example optimization in traditional MIL and give a rigorous proof. Firstly, the derivative with $\mathcal{L}$ to single instance $p^{j}$ can be calculated by:
\begin{align}
\label{eq:4}
    \frac{\partial \mathcal{L}(p)}{\partial p^{j}} &=\frac{\partial \mathcal{L}(p)}{\partial p} \frac{\partial p}{\partial p^{j}} \notag \\
    & = -\frac{1}{p} \frac{\partial p}{\partial p^{j}} \notag \\
    & = \frac{p-1}{p^{j}(1-p^{j})}
\end{align}
Similarly, for traditional MIL, we have:
\begin{align}
\label{eq:5}
    \frac{\partial \mathcal{L}(\hat{p})}{\partial p^{j}} &=\frac{\partial \mathcal{L}(\hat{p})}{\partial \hat{p}} \frac{\partial \hat{p}}{\partial p^{j}} \notag \\
    & = -\frac{1}{\hat{p}} \frac{\partial \hat{p}}{\partial p^{j}} \notag \\
    & = \frac{\hat{p}-1}{\hat{p}(1-p^{j})}
\end{align}
Note that only positive samples are considered, so here $\frac{\partial \mathcal{L}(p)}{\partial p} =- \frac{1}{p}$ and $\frac{\partial \mathcal{L}(\hat{p})}{\partial \hat{p}} =- \frac{1}{\hat{p}}$. Based on Eq.~\ref{eq:5}, we can draw that $\lim_{\hat{p}\rightarrow 1}\frac{\hat{p}-1}{\hat{p}(1-p^{j})}=0$, which explains why gradient vanishing happened in traditional MIL. 

To prove that the proposed MIL can ease the gradient vanishing problem of traditional MIL, we need to prove the following lemma:
\textbf{Lemma 1.} \textit{Given} $\mathcal{P} $ \textit{as the space spanned by} $(p^{1},p^{2},...,p^{m})$, \textit{and two subspaces} $\Omega  \subset \mathcal{P}$ \textit{and} $\hat{ \Omega } \subset \mathcal{P}$ \textit{defined as follows:}
\begin{equation}
    \Omega =\left \{(p^{1},...p^{m})|(p^{1},...p^{m})\in\mathcal{P} \ and\  \frac{\partial \mathcal{L}(p)}{\partial p^{j}} \rightarrow0\right \}
\end{equation}

\begin{equation}
    \hat{\Omega}=\left \{(p^{1},...p^{m})|(p^{1},...p^{m})\in\mathcal{P} \ and\ \frac{\partial \mathcal{L}(\hat{p})}{\partial p^{j}}
    \rightarrow0\right \}
\end{equation}
\textit{where} $m$ \textit{is the number of instances}. $\hat{\Omega}$ \textit{and} $\Omega $ \textit{satisfies:}
\begin{equation}
    \Omega \subset\hat{\Omega}
\end{equation}

We can transfer the above lemma into the proof of the following lemma:

\noindent\textbf{Lemma 2.} $\Omega $ \textit{is a proper subset of} $\hat{\Omega} $ \textit{if and only if the following two conditions are satisfied:}
\begin{align}
\label{eq:7}
& \textbf{(a)}\ \forall (p^{1},p^{2},...,p^{m}) \in \Omega,  (p^{1},p^{2},...,p^{m}) \in \hat{\Omega} \\
& \textbf{(b)}\ \exists (p^{1},p^{2},...,p^{m}) \in \hat{\Omega} , (p^{1},p^{2},...,p^{m})  \notin \Omega
\end{align}


\noindent\textbf{Proof of Lemma 2(a)} For condition (a), if $\forall (p^{1},p^{2},...,p^{m}) \in \Omega$,  $\frac{p-1}{p^{j}(1-p^{j})}\rightarrow0$, we have $p\rightarrow1$, which is equivalent to $\frac{1}{1+\prod_{j=1}^M(\frac{1}{p^{j}}-1)} \rightarrow1$. It means for each $\left \{p^{j}|j=1,2,...,m \right \}$, there exists at least one $p^{j}\rightarrow1$. Put it into the Eq.~\ref{eq:3}, we can get $\hat{p}\rightarrow1$. Then $\frac{\mathcal{L}(\hat{p})}{\partial p^{j}}=0$. Thus,  condition (a) sets up based on the above deduction.

\noindent\textbf{Proof of Lemma 2(b)} For condition (b), we need to find a special case that satisfies $\frac{\hat{p}-1}{\hat{p}(1-p^{j})}=0$ and $\frac{p-1}{p^{j}(1-p^{j})} \neq 0$. To achieve this goal, we pick two element $p^{k}, p^{q}\in \left \{p^{j}|j=1,2,...,m \right \}$ and set $p^{k}=\epsilon, p^{q}=1-\epsilon$ where $\epsilon \rightarrow 0$. Meanwhile, we set all the left $p^j$ to $\delta$ where $\delta \neq 0$ and $\delta \neq 1$.

For traditional MIL, with Eq.~\ref{eq:3}, we have $\hat{p}=1-(1-\epsilon)\epsilon\delta^{M-2} = 1$. Then $\frac{\mathcal{L}(\hat{p})}{\partial p^{j}}=\frac{\hat{p}-1}{\hat{p}(1-p^{j})}=0$, which means the gradient will vanish in this case.

As for newly proposed MIL, with Eq.~\ref{eq:1}, we have $p=\frac{1}{1+(1/\delta-1)^{M-2}}$. Then $\frac{\partial \mathcal{L}(p)}{\partial p^{j}} = {\frac{1}{1+(1/\delta-1)^{M-2}}}/{(\delta(1-\delta))} \neq 0$.


Finally, according to the proof above, gradient of the proposed MIL formulation is still nonzero while gradient vanishing happens in traditional MIL setting.

\subsection{Visualization}
We set M to 2 to show the gradient surface in 3D with respect to different inputs($p^1$ and $p^2$ are the $X$-axis and $Y$-axis respectively).
For traditional MIL, the gradient is:
\begin{align}
\label{eq:8}
    \frac{\partial \mathcal{L}(\hat{p})}{\partial p^{1}} &=\frac{\partial \mathcal{L}(\hat{p})}{\partial \hat{p}} \frac{\partial \hat{p}}{\partial p^{1}} \notag \\
    & = \frac{\hat{p}-1}{\hat{p}(1-p^{1})} \notag \\
    & = \frac{1-(1-p^1)(1-p^2)-1}{[1-(1-p^1)(1-p^2)](1-p^1)} \notag \\
    & = \frac{p^2-1}{1-(1-p^1)(1-p^2)}
\end{align}

For sharp MIL, the gradient is:
\begin{align}
\label{eq:9}
    \frac{\partial \mathcal{L}(p)}{\partial p^{1}} &=\frac{\partial \mathcal{L}(p)}{\partial p} \frac{\partial p}{\partial p^{1}} \notag \\
    & = \frac{p-1}{p^{1}(1-p^{1})} \notag \\
    & = \frac{\frac{1}{1+(\frac{1}{p^1}-1)(\frac{1}{p^2}-1)}-1}{p^{1}(1-p^{1})} \notag \\
    & = \frac{p^2-1}{p^1(2 p^1 p^2 + 1 - p^1 - p^2)}
\end{align}

As shown in Fig.~\ref{fig:contour}, the area where gradient vanishes for proposed sharp MIL is rather smaller than traditional MIL, validates that the proposed S-MIL can relieve gradient vanishing problem.

\subsection{Experiment Details}
In our experiments, dlib is adopted as the face detector for FF++ dataset, which is consistent with codes published officially by authors of Xception~\cite{dlib}. As for DFDC and Celeb datasets, we adopt MTCNN~\cite{mtcnn} with GPU for efficiency. Faces in videos are directly detected and extracted frame-by-frame with face detectors only, without any face recognition or tracking methods. All the compared methods are trained with their officially released codes and evaluated based on the above face detection results for fairness except for X-ray, who didn't release their codes and the adopted face detector, so we directly record the numbers from their original papers for fair comparison.



\end{document}


\title{Sharp Multiple Instance Learning for DeepFake Video Detection \\ (Supplementary)}
\author{Anonymous Author(s)}
\affiliation{Paper ID: 1073}

\begin{abstract}
This the supplementary of our paper sharp multiple instance learning for deepFake video detection. This document provides a detailed proof that our proposed MIL can alleviate the gradient vanishing problem existing in traditional MIL. We also provide a video on our motivation, algorithm design, and experimental application. This video is a richer demonstration of how our approach works in practice.
\end{abstract}


%

\maketitle

\section{Proof}

\begin{figure*}[t]
 \centering
  \includegraphics[width=0.8\linewidth]{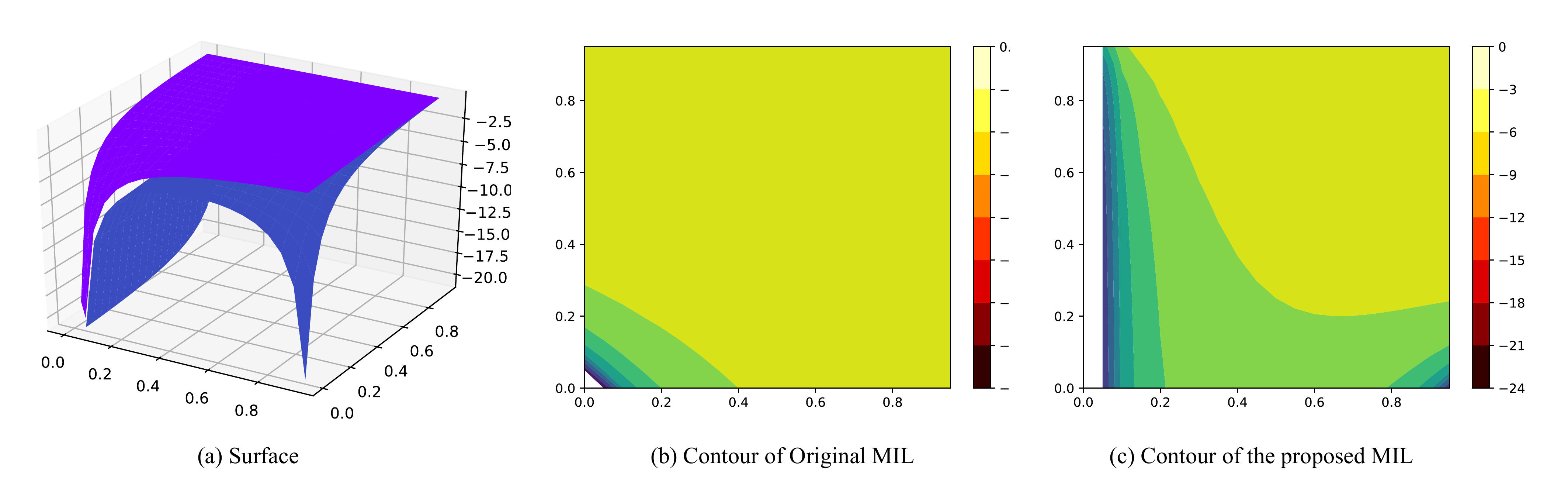}
  \caption{
          Visualization for MIL and the proposed Sharp MIL.
}
\label{fig:contour}
\end{figure*}

Section 3.2 has shown an alternative formulation for Multi Instance Learning (MIL), which is given by:
\begin{equation}
\label{eq:1}
p = \frac{1}{1+\prod_{j=1}^M(\frac{1}{p^{j}}-1)}
\end{equation}
where $p_{j}$ is the output probability of $j$-th instance and $M$ is the total number of instances. Formula~\ref{eq:1} merges the prediction of all instances and output the final probability $p$. The final training objective related to $p$ is the cross entropy loss:
\begin{equation}
\label{eq:2}
\mathcal{L}(p) = -y\log(p)-(1-y)\log(1-p)
\end{equation}
where $y$ is the ground-truth. For the convenience of derivation, we present the traditional MIL method as follows:
\begin{equation}
\label{eq:3}
\hat{p} = 1-\prod_{j=1}^M(1-p^{j}).
\end{equation}
in order to make a better distinction, we set $\hat{p}$ as the output of traditional MIL. Note that traditional MIL also has the same objective function as Eq.~\ref{eq:2}.

Next we explain why the newly proposed MIL paradigm could relieve the gradient vanishing in positive example optimization in traditional MIL and give a rigorous proof. Firstly, the derivative with $\mathcal{L}$ to single instance $p^{j}$ can be calculated by:
\begin{align}
\label{eq:4}
    \frac{\partial \mathcal{L}(p)}{\partial p^{j}} &=\frac{\partial \mathcal{L}(p)}{\partial p} \frac{\partial p}{\partial p^{j}} \notag \\
    & = -\frac{1}{p} \frac{\partial p}{\partial p^{j}} \notag \\
    & = \frac{p-1}{p^{j}(1-p^{j})}
\end{align}
similarly, for traditional MIL, we have:
\begin{align}
\label{eq:5}
    \frac{\partial \mathcal{L}(\hat{p})}{\partial p^{j}} &=\frac{\partial \mathcal{L}(\hat{p})}{\partial \hat{p}} \frac{\partial \hat{p}}{\partial p^{j}} \notag \\
    & = -\frac{1}{\hat{p}} \frac{\partial \hat{p}}{\partial p^{j}} \notag \\
    & = \frac{\hat{p}-1}{\hat{p}(1-p^{j})}
\end{align}
note that only positive samples are considered, so here $\frac{\partial \mathcal{L}(p)}{\partial p} =- \frac{1}{p}$ and $\frac{\partial \mathcal{L}(\hat{p})}{\partial \hat{p}} =- \frac{1}{\hat{p}}$. Based on Eq~\ref{eq:5}, we can draw that $\lim_{\hat{p}\rightarrow 1}\frac{\hat{p}-1}{\hat{p}(1-p^{j})}=0$, which explains why gradient vanishing happened in traditional MIL. 

\textbf{Lemma 1.} \textit{Given} $\mathcal{P} $ \textit{as the space spanned by} $(p^{1},p^{2},...,p^{m})$, \textit{and two subspaces} $\Omega  \subset \mathcal{P}$ \textit{and} $\hat{ \Omega } \subset \mathcal{P}$ \textit{defined as follows:}
\begin{equation}
    \Omega =\left \{(p^{1},...p^{m})|(p^{1},...p^{m})\in\mathcal{P} \ and\  \frac{\partial \mathcal{L}(p)}{\partial p^{j}} \rightarrow0\right \}
\end{equation}

\begin{equation}
    \hat{\Omega}=\left \{(p^{1},...p^{m})|(p^{1},...p^{m})\in\mathcal{P} \ and\ \frac{\partial \mathcal{L}(\hat{p})}{\partial p^{j}}
    \rightarrow0\right \}
\end{equation}
\textit{where} $m$ \textit{is the number of instances}. $\hat{\Omega}$ \textit{and} $\Omega $ \textit{satisfies:}
\begin{equation}
    \Omega \subset\hat{\Omega}
\end{equation}

\textbf{Lemma 2.} $\Omega $ \textit{is a proper subset of} $\hat{\Omega} $ \textit{if and only if the following two conditions are satisfied:}
\begin{align}
\label{eq:7}
& \textbf{(a)}\ \forall (p^{1},p^{2},...,p^{m}) \in \Omega,  (p^{1},p^{2},...,p^{m}) \in \hat{\Omega} \\
& \textbf{(b)}\ \exists (p^{1},p^{2},...,p^{m}) \in \hat{\Omega} , (p^{1},p^{2},...,p^{m})  \notin \Omega
\end{align}

\textbf{Proof of Lemma 1} Generally, Lemma 1 elaborates that for all value combinations $(p^{1},p^{2},...,p^{m}) \in \mathcal{P}$, the number of combinations which make $\frac{\hat{p}-1}{\hat{p}(1-p^{j})}\rightarrow 0$ is greater than which make $\frac{p-1}{p^{j}(1-p^{j})}\rightarrow0$. Lemma 1 is equivalent with two conditions stated in Lemma 2. Thus, we transform the problem to prove Lemma 2(a) and Lemma 2(b) alternatively.

\textbf{Proof of Lemma 2(a)} For condition (a), if $\forall (p^{1},p^{2},...,p^{m}) \in \Omega$, also know as $\frac{p-1}{p^{j}(1-p^{j})}\rightarrow0$ satisfied, we have $p\rightarrow1$, e.g. $\frac{1}{1+\prod_{j=1}^M(\frac{1}{p^{j}}-1)} \rightarrow1$. It means for each $\left \{p^{j}|j=1,2,...,m \right \}$, there exists least one $p^{j}\rightarrow1$. Put it into the Eq.~\ref{eq:3}, we can get $\hat{p}\rightarrow0$. Thus,  condition (a) sets up based on the above deduction.

\textbf{Proof of Lemma 2(b)} For condition (b), we need to find a special case that satisfies $\frac{\hat{p}-1}{\hat{p}(1-p^{j})}=0$ but makes $\frac{p-1}{p^{j}(1-p^{j})} \neq 0$. To achieve our goal, we pick one element $p^{k}\in \left \{p^{j}|j=1,2,...,m \right \}$ and set $p^{k}$ to 1. It ensures $\frac{\hat{p}-1}{\hat{p}(1-p^{j})}=0$. Then we set all $\frac{1}{2}$ for $\left \{p^{j}|j=1,2,...,m, j\neq k \right \}$. It makes $p^{k}=p$ satisfied. To put $p^{k}$ into $p^{j}$ of Eq.~\ref{eq:4}, we have $\frac{p-1}{p^{k}(1-p^{k})}=-\frac{1}{p^{k}}\neq0$. Condition (b) also sets up.

Finally, according to the proof above, gradient of proposed MIL formulation is still nonzero while gradient vanishing happened in traditional MIL setting. 

We set M to 2 to show the gradient surface in 3D with respect to different inputs($p^1$ and $p^2$ are the $X$-axis and $Y$-axis respectively).
For traditional MIL, the gradient is:
\begin{align}
\label{eq:8}
    \frac{\partial \mathcal{L}(\hat{p})}{\partial p^{1}} &=\frac{\partial \mathcal{L}(\hat{p})}{\partial \hat{p}} \frac{\partial \hat{p}}{\partial p^{1}} \notag \\
    & = \frac{\hat{p}-1}{\hat{p}(1-p^{1})} \notag \\
    & = \frac{1-(1-p^1)(1-p^2)-1}{[1-(1-p^1)(1-p^2)](1-p^1)} \notag \\
    & = \frac{p^2-1}{1-(1-p^1)(1-p^2)}
\end{align}

For sharp MIL, the gradient is:
\begin{align}
\label{eq:9}
    \frac{\partial \mathcal{L}(p)}{\partial p^{1}} &=\frac{\partial \mathcal{L}(p)}{\partial p} \frac{\partial p}{\partial p^{1}} \notag \\
    & = \frac{p-1}{p^{1}(1-p^{1})} \notag \\
    & = \frac{\frac{1}{1+(\frac{1}{p^1}-1)(\frac{1}{p^2}-1)}-1}{p^{1}(1-p^{1})} \notag \\
    & = \frac{p^2-1}{p^1(2 p^1 p^2 + 1 - p^1 - p^2)}
\end{align}

As shown in Fig.~\ref{fig:contour}, the area where gradient vanishes for proposed sharp MIL is rather smaller than traditional MIL, validates that the proposed S-MIL can relieve gradient vanishing problem.